\documentclass[journal]{IEEEtran}

\ifCLASSINFOpdf
\else
\fi

\usepackage[cmex10]{amsmath}
\usepackage{amssymb}
\usepackage{gensymb}
\usepackage{textcomp}
\usepackage{pifont}
\usepackage{graphics}
\usepackage{cite}
\usepackage[OT2,T1]{fontenc}
\usepackage{subfigure}
\usepackage{url}

\usepackage{float}
\usepackage{amsmath}
\usepackage{amstext}
\usepackage{amsfonts}
\usepackage[pdftex]{graphicx}
\usepackage{makecell}
\usepackage{array,multirow}
\usepackage{multicol}
\usepackage{color}
\usepackage{colortbl}

\usepackage{mathtools}
\usepackage{booktabs,tabularx}

\newcolumntype{C}[1]{>{\centering\arraybackslash}p{#1}}
\usepackage{arydshln}
\usepackage{slashbox}
\usepackage{flushend}
\usepackage{enumitem}
\usepackage[linesnumbered,ruled]{algorithm2e}

\begin{document}

%
\title{Forensicability Assessment of Questioned Images in Recapturing Detection}
%
%
%

\author{Changsheng Chen,~\IEEEmembership{Senior Member,~IEEE},
        Lin Zhao,~\IEEEmembership{Student~Member,~IEEE},
        \\Rizhao Cai,
        Zitong Yu,
        Jiwu Huang,~\IEEEmembership{Fellow, IEEE},
        Alex C. Kot,~\IEEEmembership{Fellow, IEEE}

\thanks{C. Chen, L. Zhao, and J. Huang are with the Guangdong Key Laboratory of Intelligent Information Processing and Shenzhen Key Laboratory of Media Security, and National Engineering Laboratory for Big Data System Computing Technology, College of Electronics and Information Engineering, Shenzhen University, Shenzhen, China. They are also with Shenzhen Institute of Artificial Intelligence and Robotics for Society, China (e-mail: cschen@szu.edu.cn, zhaolin2016@email.szu.edu.cn, jwhuang@szu.edu.cn).}
\thanks{R. Cai, Z. Yu and A. C. Kot are with with the Department of Electrical and Electrics Engineering, Nanyang Technological University, Singapore 639798. (e-mail: rzcai@ntu.edu.sg, zitong.yu@ntu.edu.sg, eackot@ntu.edu.sg)}
}

\maketitle

\begin{abstract}
Recapture detection of face and document images is an important forensic task.
With deep learning, the performances of face anti-spoofing (FAS) and recaptured document detection have been improved significantly.
However, the performances are not yet satisfactory on samples with weak forensic cues. 
The amount of forensic cues can be quantified to allow a reliable forensic result.
In this work, we propose a forensicability assessment network to quantify the forensicability of the questioned samples.
The low-forensicability samples are rejected before the actual recapturing detection process to improve the efficiency of recapturing detection systems.
We first extract forensicability features related to both image quality assessment and forensic tasks.
By exploiting domain knowledge of the forensic application in image quality and forensic features, we define three task-specific forensicability classes and the initialized locations in the feature space.
Based on the extracted features and the defined centers, we train the proposed forensic assessment network (FANet) with cross-entropy loss and update the centers with a momentum-based update method.
We integrate the trained FANet with practical recapturing detection schemes in face anti-spoofing (FAS) and recaptured document detection tasks.
Experimental results show that, for a generic CNN-based FAS scheme, FANet reduces the EERs from 33.75\% to 19.23\% under ROSE to IDIAP protocol by rejecting samples with the lowest 30\% forensicability scores.
The performance of FAS schemes is poor in the rejected samples, with EER as high as 56.48\%.
Similar performances in rejecting low-forensicability samples have been observed for the state-of-the-art approaches in FAS and recaptured document detection tasks.
To the best of our knowledge, this is the first work that assesses the forensicability of recaptured document images and improves the system efficiency.
\end{abstract}


\IEEEpeerreviewmaketitle
\section{Introduction}
\label{sec:Introduction}


Digital images have been used as electronic evidence in many applications, e.g., user authentication through face recognition. 
The input images with questioned authenticity need to be inspected for security in these applications.
Recapturing detection is an image forensic task that aims at distinguishing questioned images that are captured (with a single imaging process) or recaptured (with two or more imaging processes).
We can find its applications in face anti-spoofing (FAS) which classifies the genuine face images (acquired by capturing) and the spoofing face images (recaptured from printed photos, displayed videos, or face masks).
Over recent years, the performances of FAS schemes have been significantly improved by employing sophisticated deep learning networks with ever-growing complexity and supervision data with additional acquisition sensors, such as depth information \cite{li2021asymmetric}.
However, the robustness and generality of many FAS schemes are not always satisfactory, especially under challenging image distortions.
Zhang \emph{et al.} \cite{zhang2022robust} revealed that the existing methods were not working well in classifying the face images of low quality.

To investigate the generality of such observation, we have analyzed the error samples under the document recapturing detection task.
In the inter-dataset experimental protocol, the blurry and noisy document samples contribute to more than 80\% of error samples for a state-of-the-art (SOTA) scheme \cite{chen2021domain}.
We also validate the difficult samples (10 captured and 40 recaptured document images) in the document recapturing detection task with a subjective assessment experiment following the evaluation protocols in \cite{cao2010identification}. 
Our experimental results show that the average subjective classification accuracy is only 68.48\%, which is a low performance among subjects with expertise in image processing.
Therefore, image forensics under blurry and noisy samples is not a trivial task.

\begin{figure}[t!]
\centering
\centerline{\includegraphics[width=0.8\linewidth]{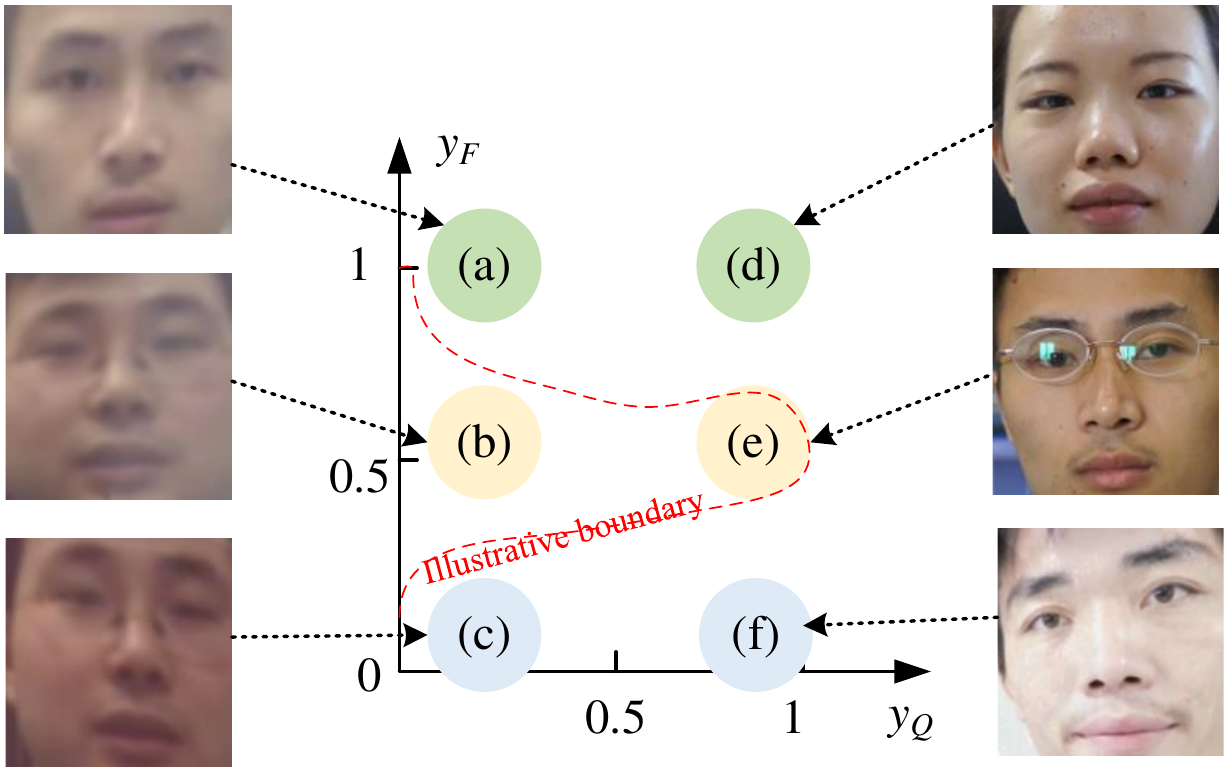}}
\caption{Examples of 6 types of representative samples in terms of forensicability features plotted in the forensicability space (Image quality score $y_Q$ versus forensic score $y_F$). Green, blue and yellow indicate genuine, spoofing and ambiguous forensic decisions, respectively. The red dashed curve shows an illustrative boundary to separate the low and high forensicability samples. (a) low $y_Q$ and high $y_F$: spoofing cues in color distortion, lighting characteristics, and face contour. (b) low $y_Q$ and ambiguous $y_F$: blurry and weak forensic trace. (c) low $y_Q$ and low $y_F$: spoofing cues in color distortion and lighting characteristics. (d) high $y_Q$ and high $y_F$: clear facial texture under natural illumination. (e) high $y_Q$ and ambiguous $y_F$: clear face image under natural illumination but with reflective glasses. (f) high $y_Q$ and low $y_F$: clear face image under unnatural illumination conditions.}\vspace{-0.25cm}
\label{fig:ForensicabilitySample}
\end{figure}


The forensic performances on these samples can rarely be improved by designing more sophisticated classifiers \cite{zhang2022robust} since the forensic cues are not sufficient to make a reliable decision.
In both FAS and the document recapturing detection tasks, the recapturing traces, such as moir\'e pattern, and color degradation have been heavily corrupted by noise and blurring distortions.
Given the above observations, we propose to first assess the forensicability of questioned images before performing image forensics to reject the samples with weak forensic cues.

In many image authentication tasks, the input images are also inspected before performing security procedures.
For example, the quality of fingerprint samples was evaluated before fingerprint authentication \cite{shen2001quality, chen2005fingerprint} to facilitate the following feature extraction and image authentication process. 
Some FAS methods employ image quality features in detecting spoofing faces \cite{li2016face, fourati2017face} with the assumption that the spoofing face images are of lower quality than the genuine ones.
Moreover, Chen \emph{et al.} \cite{chen2013two} proposed to assess the quality of 2D barcode images before the decoding process. 
This work rejects the low-quality samples in an early stage to avoid decoding errors, as well as to save processing time in the decoding and authentication steps.

However, prior works \cite{shen2001quality, chen2005fingerprint, li2016face, fourati2017face, chen2013two} only focus on the image quality features which are not sufficient in quantifying the forensicability of a sample under recapturing detection tasks.
Without loss of generality, we illustrate the differences between image quality and forensic features/scores (elaborated in Sec.~\ref{subsec:Quantification}) under the FAS task with Fig.~\ref{fig:ForensicabilitySample}.
On the one hand, the low-quality spoofing samples in (a) and (c) can be easily classified by forensic features, such as color distortion, lighting characteristics, and face contour.
On the other hand, the high-quality sample in (e) is difficult to classify due to an ambiguous forensic decision (with a forensic score near 0.5) from contradictory spoofing cues in the images, e.g., reflection on the glasses and the natural illumination.
Therefore, both image quality and forensic features are important.

In this work, we propose a forensicability assessment network (FANet) that assesses the forensicability or the amount of forensic cues in the questioned image under some representative forensic tasks, i.e., FAS and recaptured document detection. 
Our FANet consists of two steps: forensicability features extraction and forensicability quantification.
The forensicability features related to both image quality assessment and forensic tasks are employed. 
In the training phase of the forensicability quantification step, both the image quality features and forensic features of questioned images are first mapped to the corresponding quality and forensic scores with score mapping networks.
By exploiting the domain knowledge of the forensic application, we define three task-specific forensicability classes and the associated center locations in the feature space.
The weights in the score mapping networks and the locations of three class centers are updated with the stochastic gradient descent and the momentum-based update method, respectively.
In the testing phase, the quality and forensic scores of a questioned image are computed by a forward pass in the score mapping networks. 
The forensicability of the sample is then determined by the proposed forensicability distances computed from the output scores and the three class centers.

To demonstrate the efficiency of the proposed FANet, we investigate the performance of image forensicability assessment in different forensic tasks, including FAS and document recapturing detection. 
Experimental results show that, in FAS task, our FANet (only trained by synthetic data) rejects samples with low forensicability and improves the whole system performance with generic and SOTA FAS methods. 
Specifically, for a generic CNN-based FAS scheme, FANet reduces the EERs (Equal Error Rates) from 33.75\% to 19.23\% under ROSE$\rightarrow$IDIAP protocol by rejecting samples with the lowest 30\% forensicability scores.
The EER of the rejected low-forensicability samples is as high as 56.48\%.
A high EER indicates poor performance of the FAS scheme.
Similar performance gain has been observed in the task of document recapturing detection.
By rejecting the samples with the lowest 30\% forensicability scores, our FANet improves the efficiency of a SOTA FAS method by a 23.07\% improvement in EER and a 7.5\% increment in computational complexity. 

The main contributions of this work can be summarized as follows.
\begin{itemize}
    \item We propose the FANet to assess the forensicability of questioned samples by both image quality and forensic features, and quantify the forensicability scores before the actual forensic inspection. Such a forensicability assessment strategy may reject the samples with weak forensic cues or low forensicability scores, avoids making difficult forensic decisions, and improves the efficiency of the whole forensic system.
    \item We define the three forensicability classes according to the domain knowledge of a forensic task, and initialize the corresponding centers with forensic-explainable locations. The latent space of forensicability features and the center locations are updated during the training process of FANet to allow accurate characterization of the forensicability score of a questioned sample.
    \item We demonstrate under practical image recapturing detection tasks that the proposed FANet improves the efficiency of the whole forensic system. In both FAS and document recapturing detection tasks, FANet reduces over 1/3 of the classification EERs of some generic CNNs by rejecting 30\% low-forensicability samples,  with EER higher than 50\% in the rejected samples. Similar improvements have also been observed for the SOTA methods.
\end{itemize}

The remaining of this paper is organized as follows.
Section~\ref{sec:RelatedWork} reviews the related literature on forensicability under different forensic tasks and uncertainty quantification in general classification tasks.
Section~\ref{sec:Method} introduces the framework of the proposed FANet with emphasis on the selection of forensicability features, design of forensicability categories, and the algorithm of the forensicability quantification.
Section~\ref{sec:Experiment} evaluates the efficiency of the proposed FANet with both generic and SOTA forensic schemes in some practical forensic applications. 
Section~\ref{sec:Conclusion} concludes this paper.

\section{Related Work}
\label{sec:RelatedWork}

In the literature, there are some studies about the relationship between the quality metrics and authentication performances for biometric images. 
On the one hand, some works carried out different actions in the authentication process according to their quality indices. 
For example, the low-quality images can be further enhanced, pre-processed, or rejected before the authentication process \cite{shen2001quality, alonso2007comparative, alonso2011quality}. 
On the other hand, some employ image quality features in measuring the performance in biometric image authentication \cite{chen2005fingerprint, galbally2013image}, and FAS \cite{li2016face, fourati2017face, fourati2020anti}.
However, as shown in Fig.~\ref{fig:ForensicabilitySample}, the image quality metric alone is not sufficient in quantifying the forensicability of the images in the FAS problem.

There are some theoretical works on measuring the information-theoretical limitation of multimedia forensics. 
Chu \emph{et al.} \cite{chu2015information} defined forensicability as the maximum forensic information that the extracted features contain about the multimedia states, or the mutual information between features and the hypothesis of forgery operations. 
The effectiveness of this approach was demonstrated under the problem of JPEG compression by modeling the probability of perfect estimation in the maximum number of JPEG compression cycles.
Pasquini \emph{et al.} \cite{pasquini2018information} extended the Kullback–Leibler divergence measuring framework in \cite{comesana2012detection} to the detection of image resizing operation. 
The distributions of features before and after a downscaling operation with known re-scaling parameters were derived.
Chen \emph{et al.} \cite{chen2022forensicability} proposed two measurements in quantifying the forensicability of detection in the image operation chain by introducing an anti-forensic attack algorithm.
Specifically, they measured the angle and scale of the feature perturbations, as well as the mutual information scale.
The forensicability was characterized by the forensic feature variation after different image manipulation operations in an operation chain.
However, there are some limitations to the existing works.
On the one hand, \cite{chu2015information, pasquini2018information} aimed at deriving some theoretical models between forensic features and the defined forensicability metric by employing the actual forensic features in quantifying the forensicability of the targeted operations.
This is different from our framework, which aims at measuring forensicability score with some light-weighted features before the actual forensic process. 
On the other hand, \cite{chen2022forensicability} defined the forensicability metric by utilizing the original sample and the anti-forensic attack, while both are not available under the recapturing detection task. 

Low forensicability samples are those with high uncertainty in forensic classification.
Wei \emph{et al.} \cite{wei2022towards} mitigated uncertain acquisition factors in iris recognition by proposing uncertainty embedding and uncertainty-guided curriculum learning. 
The overall recognition performance was improved by an enhancement module which reduces the uncertainty from image noise and unknown acquisition conditions.
In general machine learning research, the uncertainty estimation of a generic deep learning-based classifier is an active research topic.
Deterministic uncertainty quantification (DUQ) \cite{van2020uncertainty} estimates the uncertainty of the classification results by a kernel distance between the input feature vector and the class centers.
The estimation model is trained by minimizing the sample distance to a correct center in a classification task while maximizing the distances to other incorrect centers. 
Experimental results show that DUQ performs competitively in detecting out-of-distribution samples in some dataset pairs, such as Fashion-MNIST vs. MNIST.
The following work \cite{van2021feature} further improves the uncertainty quantification performance by imposing a bi-Lipschitz constraint on the feature extractor in the context of radial basis function (RBF) networks.
However, above methods are originally designed to quantify the uncertainty in classification tasks.
The class centers are randomly initialized with Gaussian distribution.
It then measures the uncertainties of a questioned sample to the \emph{respective} classes by calculating the exponential distances from the questioned sample to the class centers.
In this work, we define the forensicability class centers with forensics-related domain knowledge and relate the distances to different centers by their forensic meaning.
Thus, the forensicability score can be computed by \emph{jointly} considering the distances to different centers.

\section{Proposed Method}
\label{sec:Method}

\begin{figure*}[t!]
\vspace{0.25cm}
\centering
\subfigure[]{
\begin{minipage}[t]{0.2\linewidth}
\includegraphics[width=1\linewidth]{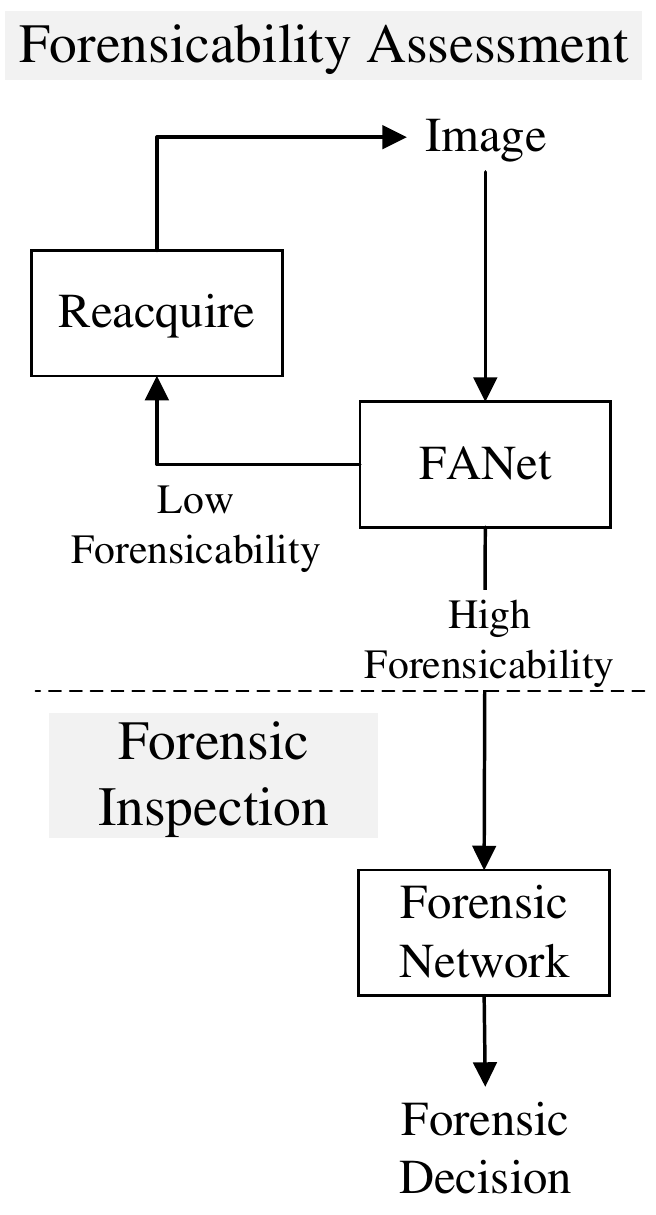}
\end{minipage}
}\hspace{1cm}
\subfigure[]{
\begin{minipage}[t]{0.7\linewidth}
\includegraphics[width=1\linewidth]{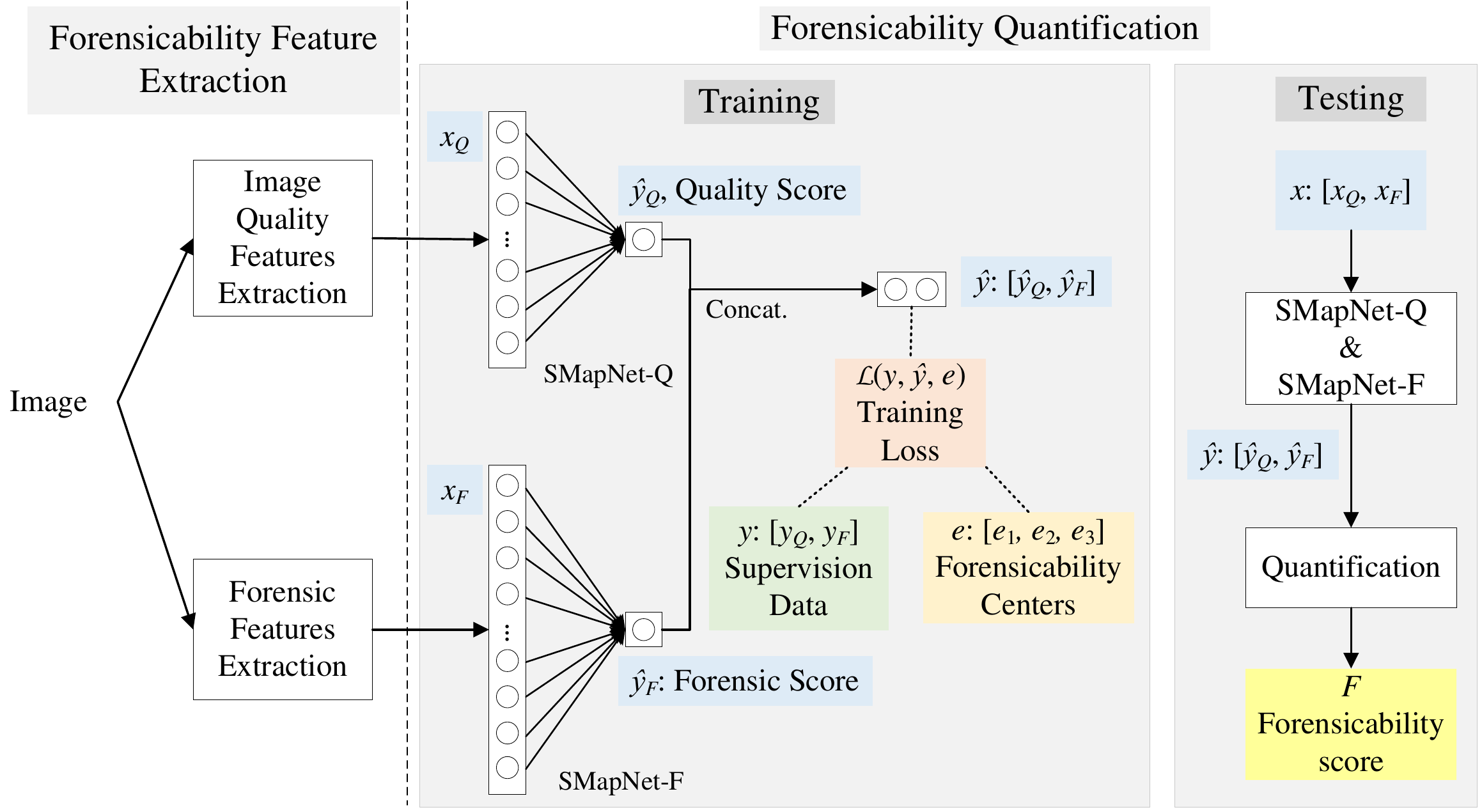}
\end{minipage}
}
\vspace{-0.25cm}
\caption{(a) The proposed two-stage forensic system, which consists of forensicability assessment and forensic inspection. (b) The proposed forensicability assessment network (FANet), including feature extraction and forensicability quantification.}\vspace{-0.25cm}
\label{fig:FANet}
\end{figure*}

In this section, we propose to assess the forensicability of questioned samples, which improves the efficiency of the authentication process.
The process is shown in Fig.~\ref{fig:FANet}~(a). 
For a questioned image, its forensicability is first evaluated by the proposed Forensicability Assessment Network (FANet). 
If the forensicability score is lower than a predetermined threshold, the system rejects the image without going through the next stage, and prompts the user for image reacquisition. 
Only when the forensicability score is above the threshold, the forensic inspection process is performed. 
As shown in Fig.~\ref{fig:FANet}~(b), FANet consists of feature extraction and forensicability quantification. 
The feature extraction step extracts forensicability features from the questioned image.
The forensicability quantification step maps the extracted features to forensicability score.



\subsection{Forensicability and its Features}
\label{subsec:Features}

In order to assess the forensicability of an image, we select suitable features with two criteria. 
First, the selected features should have a strong correlation with the forensicability of an image. 
Second, the computational complexity of feature extraction must be low since forensicability assessment is a preprocessing step in a forensic system.
Without loss of generality, we employ face anti-spoofing (FAS) as a practical forensic task to illustrate the design of the proposed FANet.
Considering the above factors, we define \textbf{forensicability} as \emph{the amount of forensic cues characterized by image quality features and light-weighted FAS features}.

$\bullet$ \emph{Image Quality Features}: Image quality features have been employed in biometric authentication systems to reject low-quality and unreliable samples \cite{alonso2007comparative, alonso2011quality}. 
Thus, we believe that image quality can also characterize the forensicability of a face image or other images to be inspected.
We employ the no-reference image quality assessment methods to extract image quality features since there is rarely a reference image in practical forensic tasks.
Inspired by \cite{fourati2020anti}, we choose three no-reference image quality assessment methods to obtain image quality features, which are BIQI \cite{moorthy2010two}, BRISQUE \cite{mittal2012no}, and GM-LOG-BIQA \cite{xue2014blind}. It is shown in \cite{fourati2020anti} that these features have good performance in face IQA tasks and low computational complexity. 
The metrics BIQI, GM-LOG-BIQA, and BRISQUE yield 18, 36, and 40 feature values, respectively, for each image sample. 
In total, a 94-dimensional feature is obtained to evaluate the image quality of an image in the recaptured image detection task.





\begin{figure}[t!]
\vspace{0.25cm}
\centerline{\includegraphics[width=0.9\linewidth]{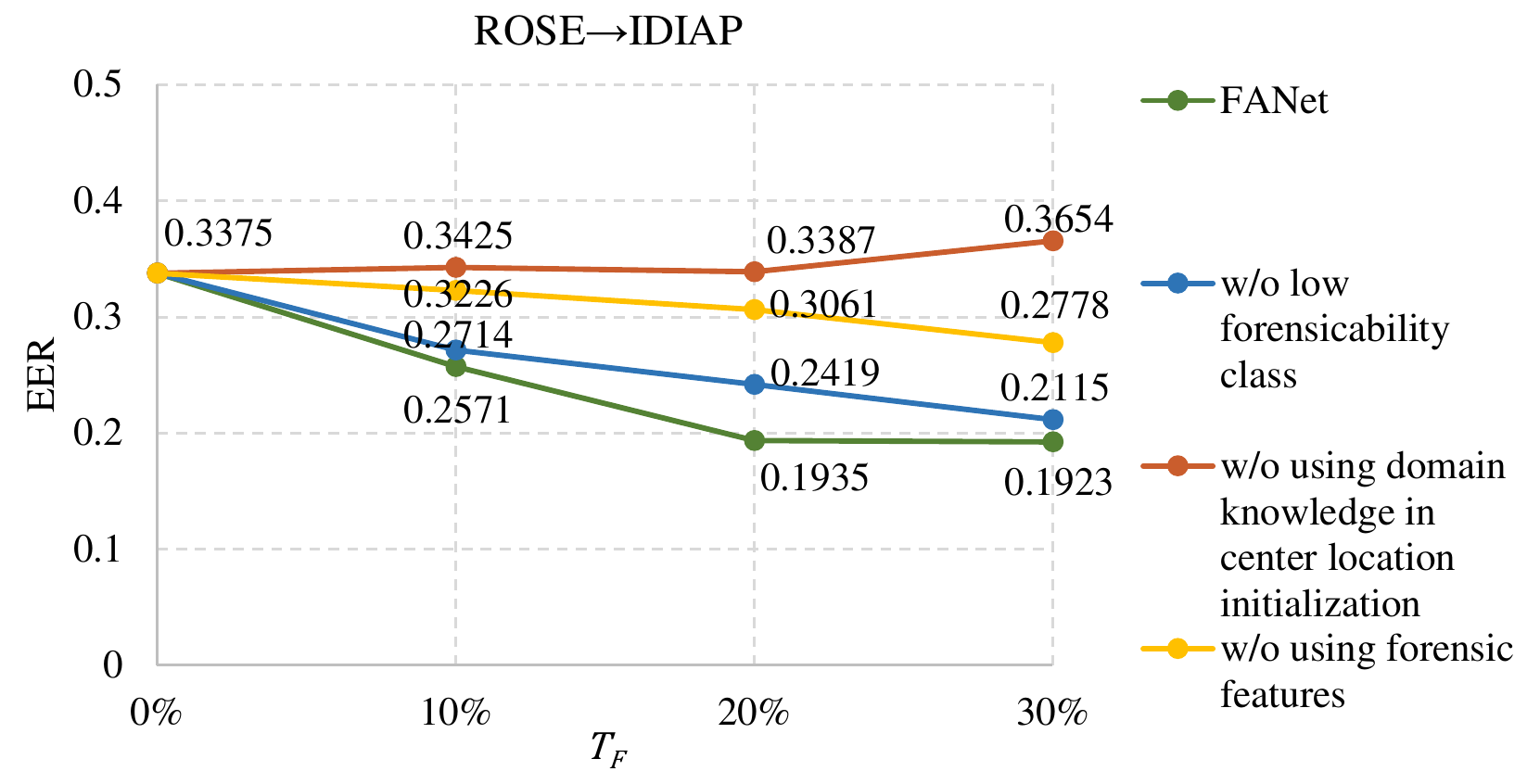}}
\caption{The ablation study of four different forensicability assessment methods: 1) our FANet, 2) w/o low forensicability class (with only the high and low forensicability classes), 3) w/o using forensic domain knowledge in center location initialization (with random initialized center locations), and 4) w/o using forensic features (with image quality features alone).}\vspace{-0.25cm}
\label{fig:FaceEERAblation}
\end{figure}

$\bullet$ \emph{Forensic Features}: As demonstrated by the illustrative examples in Fig.~\ref{fig:ForensicabilitySample} and the ablation study in Fig.~\ref{fig:FaceEERAblation}, image quality features alone are not sufficient to assess the forensicability of images.
Since we aim at measuring the forensic cues from some recaptured images, the extracted features should characterize the global forensic response.
To avoid large computational overhead, a light-weighted CNN can be employed as the forensic feature extractor.
Referring to the method in \cite{yang2014learn}, we use an 8-layer CNN end-to-end feature extractor for mining forensic clues. 
To balance the dimensions of image quality features and forensic features, we add a 128-dimensional fully-connected (FC) layer behind the original 4096-dimensional FC layer. 
The feature extractor finally yields a feature vector of 128-dimension.






Finally, we obtain the forensicability features, including 94-dimensional image quality features and 128-dimensional forensic features.
To this end, we have extracted the image quality and forensic features in describing the forensicability of the questioned images.
Though these features are much simpler than many data-driven features in SOTA forensic methods, they are capable in the forensicability assessment. 
In other words, we believe that evaluating the strength of forensic cues in the question images is much easier than performing actual forensics.

\subsection{Forensicability Quantification}
\label{subsec:Quantification}





As shown in Fig.~\ref{fig:FANet}~(b), we quantify the image quality features and forensic features into a forensicability score. 
This process can be completed in three steps.

First, we obtain the supervision data, i.e., ground-truth quality and forensic scores, which will be employed in training our score mapping networks. 
These networks map the image quality and forensic features to image quality and forensic scores, respectively. 
The ground-truth quality scores of our samples are generated from a classifier that is pretrained by the mean opinion scores in the LIVE database \cite{sheikh2006statistical}.
The quality score is denoted as $y_Q$ and is normalized to $[0,1]$, where 0 and 1 represent low and high-quality samples.
To generate the forensic scores, we connect an FC layer behind the CNN feature extractor \cite{zhang2012face}. 
This CNN is pre-trained on the training set of the CASIA FASD database (to be covered in Sec.~\ref{subsec:Databases}). 
The FC layer maps forensic features to classification probabilities, i.e., the ground-truth forensic scores in the proposed framework. 
The forensic scores, denoted as $y_F$, are also normalized to $[0, 1]$, where 0 and 1 indicate spoofing and genuine images. 
If the forensic score is close to 0.5, it means that the FAS classifier has difficulty distinguishing the class of the questioned image. 
To this end, the supervision information of image quality and forensics are obtained as $y_Q$ and $y_F$.
Fig.~\ref{fig:ForensicabilitySample} visualizes 6 types of representative samples in a 2D feature space with image quality score $y_Q$ versus forensic score $y_F$.


Second, the scores $y_Q$ and $y_F$ are employed in training our score mapping networks SMapNet-Q and SMapNet-F which map the image quality features $x_Q$ and forensic features $x_F$ to image quality score $\hat{y}_Q$ and forensic score $\hat{y}_F$, respectively. 
During the training process, we exploit the domain knowledge of forensic tasks and define three forensicability categories (high positive, high negative, and low forensicability) and the associated center locations in the feature space of image quality versus forensic.
The class labels of training samples are then determined by a customized distance toward three forensicability centers.
By employing the supervision data (i.e., scores $y_Q, y_F$), the output from score mapping networks $\hat{y}_Q, \hat{y}_F$, and the corresponding class labels, a loss function is built by the sum of the binary cross-entropy losses in the three categories.
In the training process, the weights in score mapping networks can be updated by stochastic gradient descent with our loss function, while the locations of three class centers are updated by a momentum-based update method.

Third, we perform forensicability quantification on the quality and forensic scores from questioned samples. 
In the second step, our score mapping networks are trained, and the output quality and forensic scores of testing samples, denoted as $\hat{y}_Q$ and $\hat{y}_F$ can be generated.
The distance between the score of testing samples $\hat{y}: [\hat{y}_Q, \hat{y}_F]$ and three class centers are computed accordingly.
The forensicability score of a questioned sample is determined by the proposed forensicability distances which is a weighted combination of linear and exponential distances between sample score and three class centers.

\subsection{Design Details}
\label{subsec:DesignDetails}

In the following parts, we will elaborate on some important details in the above three steps presented in Sec.~\ref{subsec:Quantification} .




\subsubsection{Supervision Data: Definition of Forensicability Centers}
\label{subsubsec:SupervisionData}

As shown in Fig.~\ref{fig:ForensicabilitySample}, the forensicability of images shows a huge difference in terms of image quality and forensic cues. 
Therefore, in the forensicability assessment task, images can also be classified according to their forensicability, i.e., high and low forensicability. 
Considering that the genuine forensic features and spoofing forensic features are significantly different, we increase the number of forensicability classes to three classes, defining high image quality and strong positive (genuine) forensic cues (short for high positive forensicability), high image quality and strong negative (spoofing) forensic cues (short for high negative forensicability), and low image quality and weak forensic cues (short for low forensicability).

Based on the supervision information $y: [y_Q, y_{F}]$, we define three forensicability centers by exploiting intuition from the forensic task. 
\begin{enumerate}
    \item $e_1$: [1.0, 1.0], center for high positive forensicability;
    \item $e_2$: [1.0, 0.0], center for high negative forensicability;
    \item $e_3$: [0.0, 0.5], center for low forensicability. 
\end{enumerate}

\begin{figure}[t!]
\centering
\subfigure[]{
\begin{minipage}[t]{0.45\linewidth}
\centering
\includegraphics[width=1.5in]{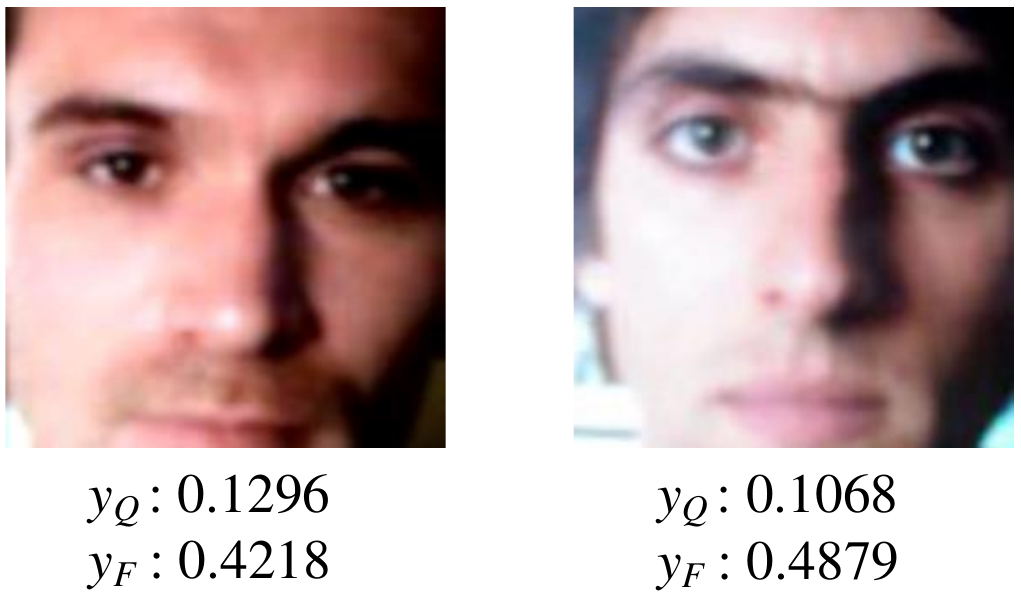}
\end{minipage}
}
\enspace
\subfigure[]{
\begin{minipage}[t]{0.45\linewidth}
\centering
\includegraphics[width=1.5in]{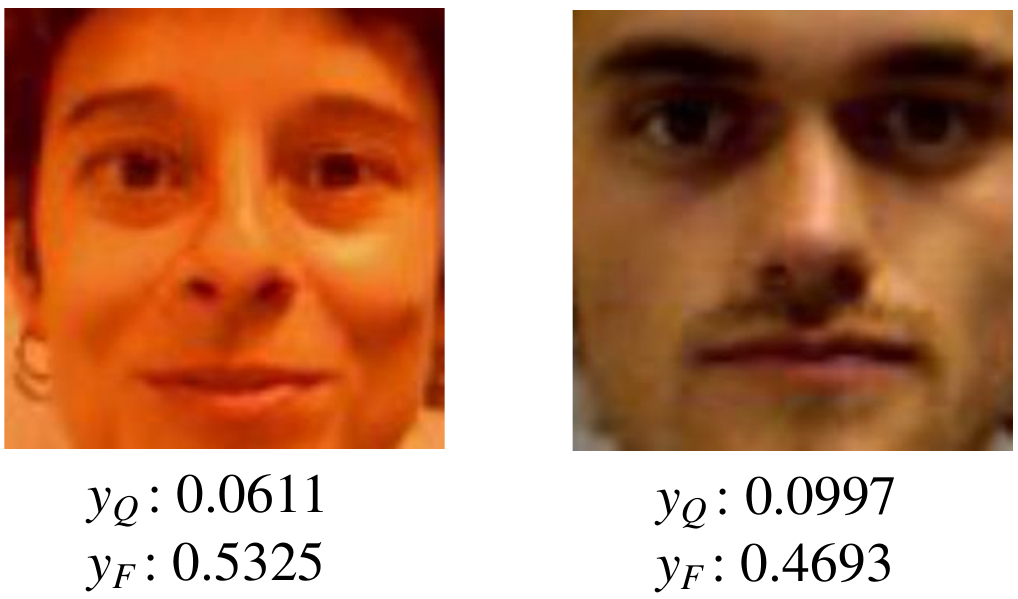}
\end{minipage}
}\vspace{-0.25cm}
\caption{Some low-forensicability samples from IDIAP testing set. $y_Q$ denotes image quality score and $y_{F}$ denotes forensic score. (a) Spoofing faces. (b) Genuine faces.}\vspace{-0.25cm}
\label{fig:FaceLowForensicability}
\end{figure}

In the training phase, the samples are classified according to the distance between the supervised data $y$ and the class centers. 
The distance from the $i$-th sample (with supervised data $y_i: [y_Q, y_F]_i$) to the $c$-th center is 
\begin{align}
\label{eqn:EuclideanDistance}
D_i^c = \mathcal{D}(y_i, e_c),
\end{align}
where $\mathcal{D}(\cdot)$ denotes the Euclidean distance. 
The training set is first sorted by $\mathcal{D}(y_i, e_3)$ in descending order. 
During the training process, 30\% samples with the shortest distances are labeled as samples with low forensicability. 
Selecting 30\% samples as the low forensicability samples leads to a balanced number of samples in the three categories during the training period.
Moreover, as illustrated in Fig.~\ref{fig:FaceLowForensicability}, the selected low forensicability samples in current setting are difficult to classify even by humans.
The remaining samples are classified by comparing $\mathcal{D}(y_i, e_1)$ and $\mathcal{D}(y_i, e_2)$. 
If $\mathcal{D}(y_i, e_1)>\mathcal{D}(y_i, e_2)$, a sample is classified as high positive forensicability.
Otherwise, it is of high negative forensicability.


\subsubsection{Training Process of Forensicability Assessment Network}



In the previous parts, we have obtained the supervision data $y: [y_Q, y_F]$ from the training samples and have defined the associated labels, i.e., forensicability categories.
Here, we present the details of the training process of our FANet (the middle gray parts in Fig.~\ref{fig:FANet}~(b)) with the pseudo-code provided in Algorithm~\ref{algo:FANet}. Only the weights in SMapNet-Q and SMapNet-F and the following layers are updated during the training of FANet.

To facilitate the training, we define the loss function as the sum of the binary cross-entropy losses in the corresponding category, which is 
\begin{align}
\label{eqn:Loss}
\mathcal{L}(y_i, \hat{y}_i, e_c) = -\sum_{c=1}^3 {y}_i^c \log(K_i^c) + (1 - {y}_i^c)\log(1 - K_i^c)
\end{align} 
where $y_i^c$ is binary indicator for the $c$-th class with $y_i^c=1$ indicates $y_i$ is labeled as the $c$-th class.
$K_i^c$ is the RBF kernel function of sample score $\hat{y}_i$ to each class center, which is defined as
\begin{align}
\label{eqn:RBFKernel}
K_i^c = \exp{\left[-\frac{\| \lambda_c (\hat{y}_i - e_c) \|_2^2}{2\sigma^2} \right]},
\end{align}
where the operation $\| \cdot \|_2$ takes the L2 norm. 
The parameter $\sigma$ denotes the RBF kernel size, which is empirically set to $\sigma=0.1$. 
The parameter $\lambda_c$ is a trainable 2-dimensional weights for balancing image quality and forensics scores.
According to the definition in Eq.~(\ref{eqn:RBFKernel}), $K_i^c$ is an exponentially scaled metric in $[0,1]$, which can then be employed in computing the cross-entropy loss in Eq.~(\ref{eqn:Loss}).

\begin{algorithm}[t!]
    \caption{Pseudo-code for training FANet}
    \label{algo:FANet}
    \SetKwInOut{Input}{Input}
    \SetKwInOut{Output}{Output}

    \Input{- Forensicability features $x: [x_Q, x_F]$;\\
           - Supervised information $y: [y_Q, y_F]$;\\
           - Forensicability centers $e_c$;\\
           - RBF kernel size $\sigma$.
    }

    Compute $D_i^c$ according to Eq.~(\ref{eqn:EuclideanDistance}).

    Determine the class labels $y_i^c$ by $D_i^c$.

    \textbf{While} (not at end of epochs) \textbf{do}

        \hspace{0.5cm}\hangindent=0.5cm Compute the output $\hat{y}_Q$ and $\hat{y}_F$ by forward propagation of SMapNet-Q and SMapNet-F.
        
        \hspace{0.5cm}\hangindent=0.5cm Concatenate $\hat{y}_Q$ and $\hat{y}_F$ to $\hat{y}_i: [\hat{y}_Q, \hat{y}_F]_i$.
        
        \hspace{0.5cm}\hangindent=0.5cm Compute the RBF kernel function $K_i^c$ by Eq.~(\ref{eqn:RBFKernel}).

        \hspace{0.5cm}\hangindent=0.5cm Compute the loss $\mathcal{L}(y_i, \hat{y}_i, e_c)$ according to Eq.~(\ref{eqn:Loss}).
        
        \hspace{0.5cm}\hangindent=0.5cm Compute the gradient of neurons according to loss. 
        
        \hspace{0.5cm}\hangindent=0.5cm Back propagation to update $\lambda_c$ and the weights of SMapNet-Q and SMapNet-F.
        
        \hspace{0.5cm}\hangindent=0.5cm Update the locations of centers $e_c$ by Eq.~(\ref{eqn:Update}).

    \textbf{End}

    \Output{FANet with trained weights and centers.}
\end{algorithm}

During the training process, we minimize the loss function and perform stochastic gradient descent on the network parameters as well as $\lambda_c$.
After backward propagation, the location of $e_c$ is updated with the scores $\hat{y}_i$ of samples in the $c$-th class.
The location of center $e_c$ in current mini-batch is computed following \cite{van2017neural} as 
\begin{align}
\label{eqn:Update}
e_{c} &= \frac{\eta \cdot \nu_{c}' + (1 - \eta) \cdot \sum_{i \in \mathbb{C}} \hat{y}_{i}}{\eta \cdot N_{c}' + (1 - \eta) \cdot N_c},
it\end{align}
where $\eta$ is the momentum, which we set to 0.9.
$N_c$ is the number of samples belonging to the $c$-th class within the current mini-batch, and $N_{c}'$ is the result of the last update.
$\nu_{c}'$ is the update direction in the last mini-batch.
It is computed by the difference of center locations in the current mini-batch and the last update.
Set $\mathbb{C}$ contains all samples with class label $c$ in the current mini-batch.
Therefore, $\sum_{i \in \mathbb{C}} \hat{y}_{i}$ is the sum of all sample locations in the weighted feature space of the $c$-th class in the current mini-batch.
The momentum-based class center updating method reduces the sensitivity of the update direction of features in a particular batch, thus speeding up the training process.

\subsubsection{Testing Process of Forensicability Assessment Network}

During the testing (the right gray block in Fig.~\ref{fig:FANet}~(b)), we define the forensicability score of a sample by
\begin{align}
\label{eqn:Forensicability}
F_i = \beta (1-K_i^3) + (1 - \beta) \| \hat{D}_i^1 - \hat{D}_i^2 \|_1,
\end{align}
where $ \hat{D}_i^1$ and $\hat{D}_i^2$ are the Euclidean distances from the output $\hat{y}_i$ to the high positive and high negative forensicability centers ($e_1$ and $e_2$), $\| \cdot \|_1$ computes the L1 norm.
In the first term, $(1-K_i^3)$ measures the kernel distance between $\hat{y}_i$ and the low-forensicability center $e_3$. 
The second term computes the distance difference between $\hat{y}_i$ to $e_1$ and $e_2$. 
Variable $\beta$ is the weights of two distances. 
It is set to 0.5 in our experiments.
Overall, a small forensicability score $F_i$ indicates a low forensicability sample.
It is noted that, according to Eq.~(\ref{eqn:RBFKernel}), the exponential function in $K_i^3$ imposes heavy weights on samples near the low-forensicability center $e_3$.
This is because during the generation of our supervision data (in Sec.~\ref{subsubsec:SupervisionData}), $\mathcal{D}(y_i, e_3)$ is employed as the first criteria in the selection of low-forensicability samples.
Moreover, our results in Sec.~\ref{subsec:Novelties} show that the exponential function in $K_i^3$ leads to more intuitive feature spaces of the low-forensicability samples.

\subsection{Analysis of Forensicability Assessment Network}
\label{subsec:Novelties}


In this part, we analyze our FANet by contrasting it with the designs in existing work on uncertainty estimation in a machine learning problem, such as DUQ \cite{van2020uncertainty}.

\begin{figure}[t!]
\centering
\subfigure[]{
\begin{minipage}[t]{0.34\linewidth}
\includegraphics[width=1\linewidth]{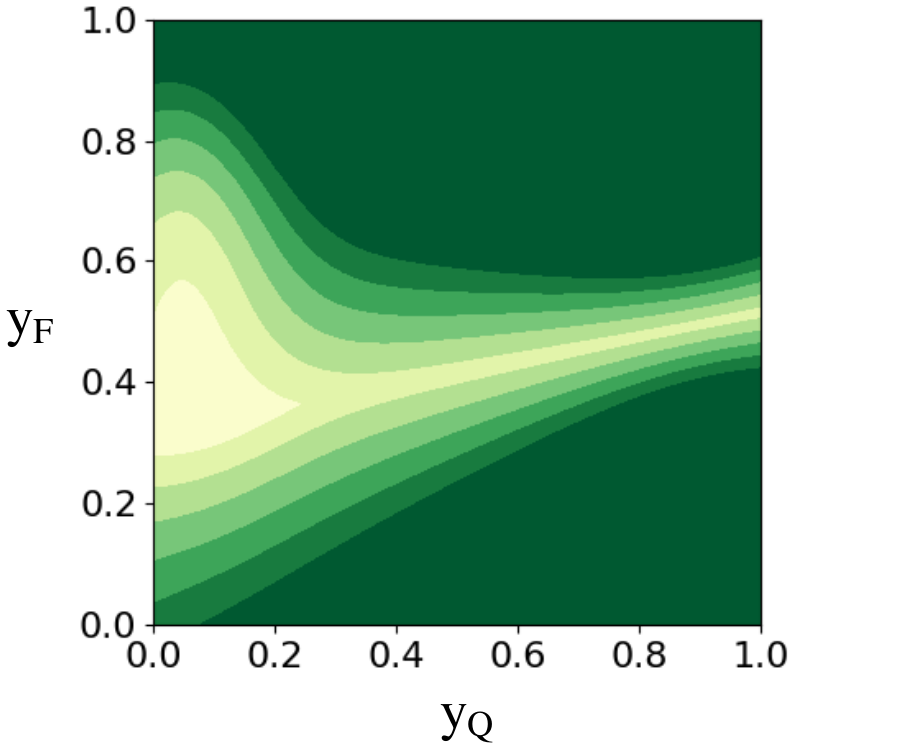}
\end{minipage}
}\hspace{-0.7cm}
\subfigure[]{
\begin{minipage}[t]{0.34\linewidth}
\includegraphics[width=1\linewidth]{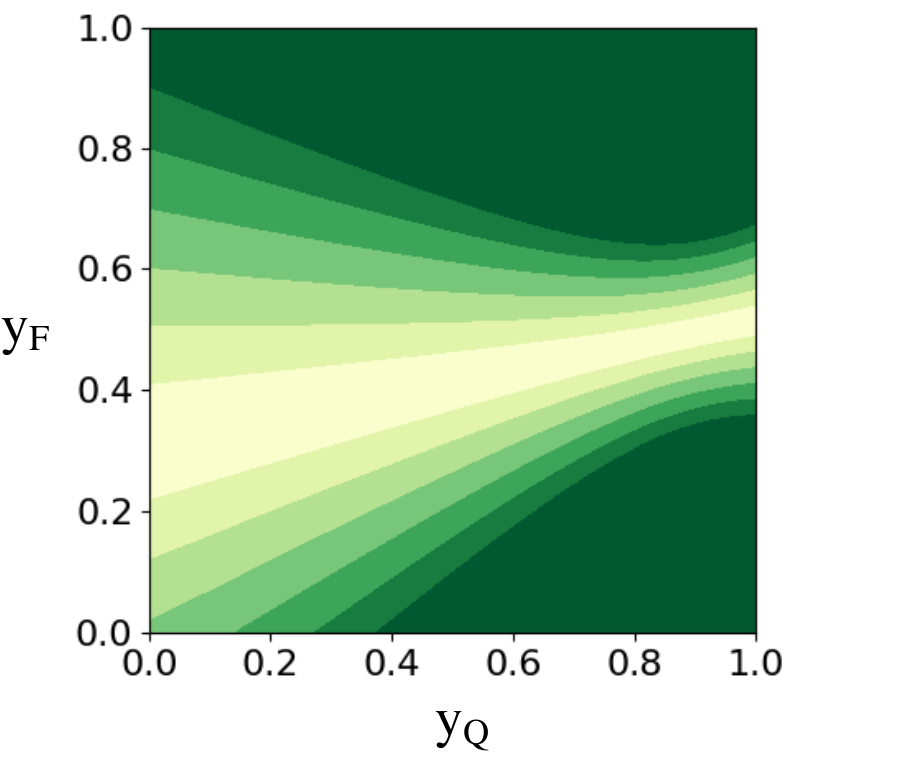}
\end{minipage}
}\hspace{-0.7cm}
\subfigure[]{
\begin{minipage}[t]{0.34\linewidth}
\includegraphics[width=1\linewidth]{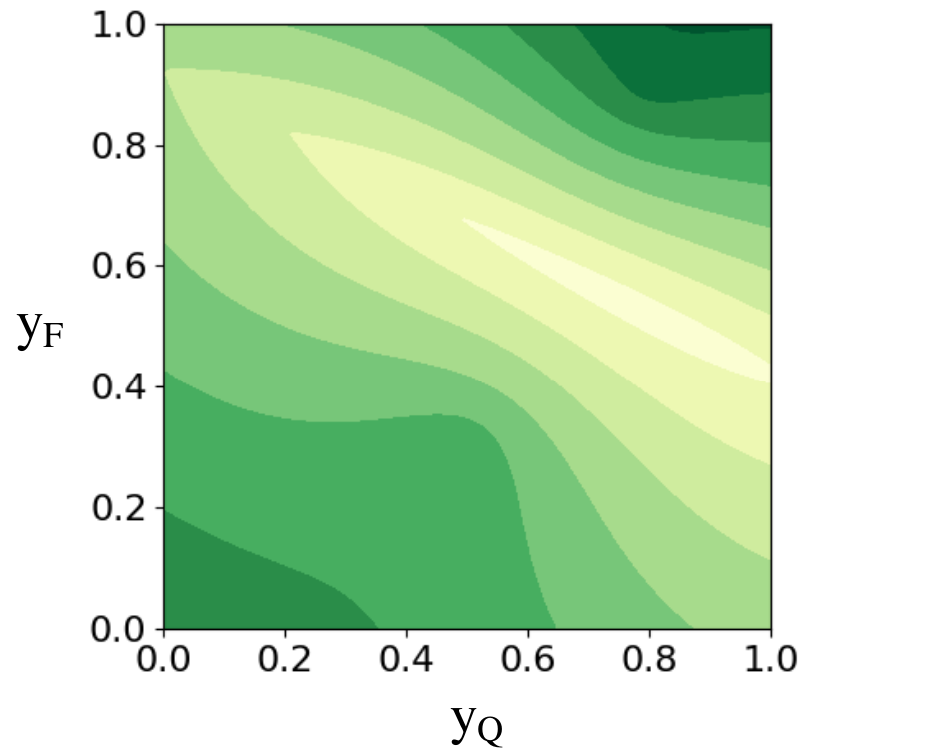}
\end{minipage}
}\vspace{-0.25cm}
\caption{Ablation study of the forensicability quantification method. The figures illustrate the distributions of forensicability scores in the quality versus forensic latent space under different configurations of class centers. A bright intensity indicates a low forensicability score. (a) The proposed forensicability quantification method. (b) w/o low forensicability class. (c) w/o using domain knowledge in center location initialization.}\vspace{-0.25cm}
\label{fig:ForensicabilityDistribution}
\end{figure}

The first difference in our method is in the setting of the number of forensicability centers. 
The existing literature, such as DUQ \cite{van2020uncertainty}, sets the number of class centers according to the number of categories in the classification task.
In FANet, we first set two classes (i.e., the high positive and the high negative forensicability classes) according to the output of the forensic application, and introduce a low forensicability class to cover the samples with weak forensic cues.
We design an ablation study on the latent space of synthetic datasets generated by the CASIA database (to be covered in Sec.~\ref{subsec:Databases}) to illustrate the importance of the low forensicability class.
A bright intensity plotted in the feature space indicates the region with low forensicability scores. 
Ideally, the high and low intensity boundary should separate the low and high forensicability samples in Fig.~\ref{fig:ForensicabilitySample}.
As shown in Fig.~\ref{fig:ForensicabilityDistribution}~(a), with the low forensicability class, the class center of the low forensicability class corrects for the distribution of low-forensicability regions, making the forensic quantification more accurate.
In contrast, without the low forensicability class, the low-forensicability regions in Fig.~\ref{fig:ForensicabilityDistribution}~(b) are distributed at the boundary of the two strong-forensicability classes.
The poor performance is mainly due to the fact that the low forensicability samples were not employed in the training process and the forensicability scores are computed without the low forensicability center.
Therefore, our setting of three class centers leads to a better performance in identifying the low forensicability samples.


The second difference is in the initialized locations of the three class centers.
For many deep learning models,  it is often difficult to relate points in the latent space with practical meanings. 
Therefore, DUQ \cite{van2020uncertainty} sets the locations of class centers by Gaussian random initialization.
In FANet, we propose to first map the image quality and forensic features to interpretable scores.
The locations of class centers are then initialized with different image quality and forensic scores according to the domain knowledge in a forensic task, i.e., centers for high positive forensicability, high negative forensicability and low forensicability.
Moreover, we illustrate why precise initialization of class centers is important in FANet by an ablation study.
As shown in Fig.~\ref{fig:ForensicabilityDistribution} (c), the low-forensicability regions are not accurate with center locations that are randomly initialized from a Gaussian distribution.
The low-forensicability samples shown in Fig.~\ref{fig:ForensicabilitySample} can not be identified by the bounardy in Fig.~\ref{fig:ForensicabilityDistribution} (c).
Thus, our setting on locations of three forensicability centers results in better performance.

\section{Experimental Results}
\label{sec:Experiment}

In this section, we investigate the performances of image forensicability assessment in two representative forensic tasks, i.e., FAS and recaptured document detection.

\subsection{Databases}
\label{subsec:Databases}


In our experiments, we utilize four publicly available face presentation attack databases, including CASIA Face Anti-Spoofing Database \cite{zhang2012face}, IDIAP REPLAY-ATTACK \cite{chingovska2012effectiveness}, the Spoofing in the Wild (SiW) database \cite{liu2018learning} and the ROSE-YOUTU database \cite{li2018unsupervised}. 
In addition, we also utilize a recaptured document image database \cite{chen2021domain} to evaluate the generalization performance of our proposed method in different forensic tasks.

\subsubsection{CASIA-FASD}
\label{subsubsec:CASIA}

The CASIA Face Anti-Spoofing Database \cite{zhang2012face} includes attack videos from print attacks (flat, wrapped, cut) and replay attack (tablet). 
In our experiments, the training set of CASIA-FASD database is used to train forensicability features, which includes the image quality assessment features and forensic features. 
The testing set is used to generate a batch of synthetic datasets for training the FANet. 
In this way, we demonstrate that our method performs well even when no spoofing data are available.

\subsubsection{IDIAP REPLAY-ATTACK}
\label{subsubsec:IDIAP}

The IDIAP REPLAY ATTACK database \cite{chingovska2012effectiveness} consists of two types of spoofing videos, i.e., print attack (flat) and replay attacks (tablet, phone) under both fluorescent lamp and daylight lighting conditions.

\subsubsection{SiW}
\label{subsubsec:SiW}

The Spoofing in the Wild database\cite{liu2018learning} covers spoofing videos of print attacks (flat, wrapped) and replay attacks (phone, tablet, monitor).

\subsubsection{ROSE-YOUTU}
\label{subsubsec:ROSE}

The ROSE-YOUTU Face Liveness Detection Database \cite{li2018unsupervised} contains three types of spoofing video, i.e., print attack (flat), replay attacks (monitor, laptop) and mask attacks (paper, crop-paper).
The data is diverse, covering up to 5 front-facing phone cameras and 5 different illumination conditions. 

\subsubsection{Document Image Database}
\label{subsubsec:Document Image Database}

The document image database consisting captured and recaptured document images was proposed in \cite{chen2021domain} for the problem of document recapturing detection. 
Student ID cards from 5 different universities are selected by the database as document templates and captured and recaptured by different printers, scanners and mobile phones. 
In our experiments, there are two stages, forensicability assessment and forensics, respectively. 
To allow fair comparison, two different datasets are used to train the model in both stages, and a third dataset is required for cross-dataset testing. 
To facilitate our evaluation, we reorganize the database in \cite{chen2021domain} into 3 datasets, $D_0$, $D_1$ and $D_2$ such that the devices used in 3 datasets are different. 
After splitting, $D_0$ contains 24 genuine and 48 recaptured samples, $D_1$ contains 60 genuine and 300 recaptured samples, and $D_2$ contains 48 genuine and 384 recaptured samples.

In summary, for the face images, we utilize the CASIA-FASD database and some generated samples for training and testing of FANet. 
The IDIAP REPLAY-ATTACK database, SiW dababase and ROSE database are employed for cross-database experiments of forensic networks. 
Similarly, for the document images, we utilize $D_0$ dataset for training and testing of FANet, and $D_1$ dataset and $D_2$ dataset for cross-database experiments of forensic networks.


\subsection{Experiments Settings}
\label{subsec:Settings}

\subsubsection{Implementations}
\label{subsubsec:Implementations}

The experiments are implemented in two parts, i.e., the training of the FANet and the application of FANet in practical forensic tasks.

\vspace{0.25cm}
\noindent $\bullet$ \emph{Training of FANet}

\begin{figure}[t!]
\centerline{\includegraphics[width=3.5in]{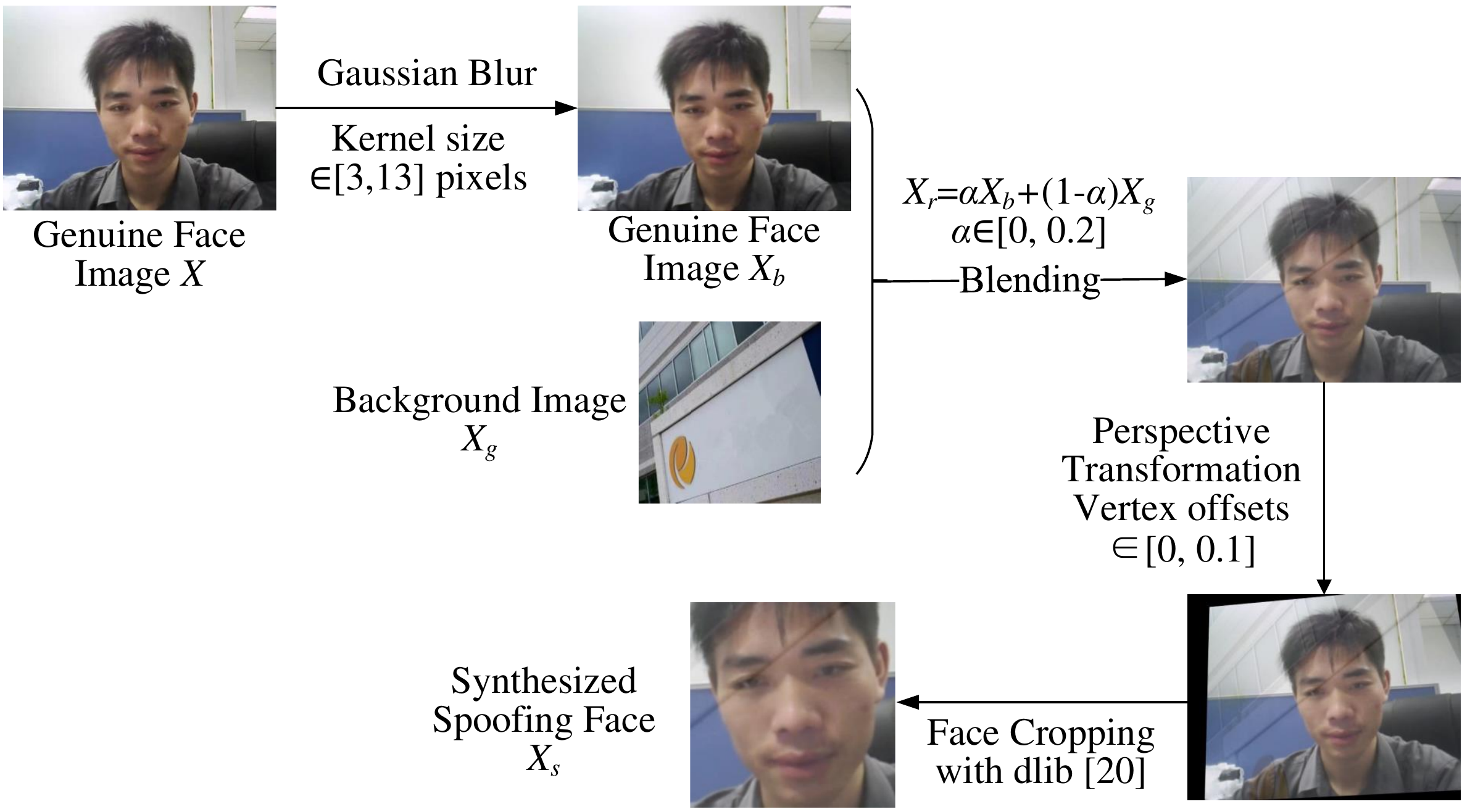}}
\caption{The process of synthesizing spoofing face images. A genuine face image is Gaussian blurred, blended with a background image, perspective transformed, and cropped to get a synthetic spoofing face image.}\vspace{-0.25cm}
\label{fig:DatabaseSynthesis}
\end{figure}

Inspired by \cite{yang2019face}, we produce a large amount of new samples by an image synthesis method, and employ the generated data to train FANet in our experiments. 
The process of generating datasets is shown in Fig.~\ref{fig:DatabaseSynthesis}. 
For FAS problem, we select genuine face images in the CASIA database as input for the data synthesis method and generate more than 14K synthetic face images for training FANet.
Here, we demonstrate to train the FANet with low-cost samples while achieving high performance in identifying low-forensicability samples.
If more diverse samples are employed in training the FANet, the performance can be further improved.
For the problem recaptured document detection, data augmentation process similar to that in Fig.~\ref{fig:DatabaseSynthesis} (except that no face cropping step is needed for document images) is performed with the document images in $D_0$. 
The images are divided into patches, which are augmented by 10 different parameter configurations to yield about 30K patches for training FANet.

Regarding the training setup of the forensicability assessment model, we use the ReLU activation function and the SGD optimizer with learning rate 0.01 (decayed by a factor of 5 every 10 epochs), momentum 0.9, weight decay rate $10^{-4}$. 
The batch size is set to 128 and the training process iterates for 50 epochs. 
The center locations are updated with $\gamma=0.999$. 
The input dimension of the model is 222 and the output dimension is 2, with a fully connected layer of 2 hidden units connected in between. 
The dimensions of the 3 centers are kept the same with the output dimension and are initialized to $[1, 0]$, $[1, 1]$ and $[0, 0.5]$, respectively, as defined in Sec.~\ref{subsubsec:SupervisionData}.
Our implementation is based on Pytorch 1.6.0 and an Nvidia Tesla V100 GPU.

\vspace{0.25cm}
\noindent $\bullet$ \emph{Application of FANet}

\begin{figure*}[t!]
\subfigure[HQ→LQ]{
\begin{minipage}[t]{\textwidth}
\centering
\includegraphics[width=2.5in]{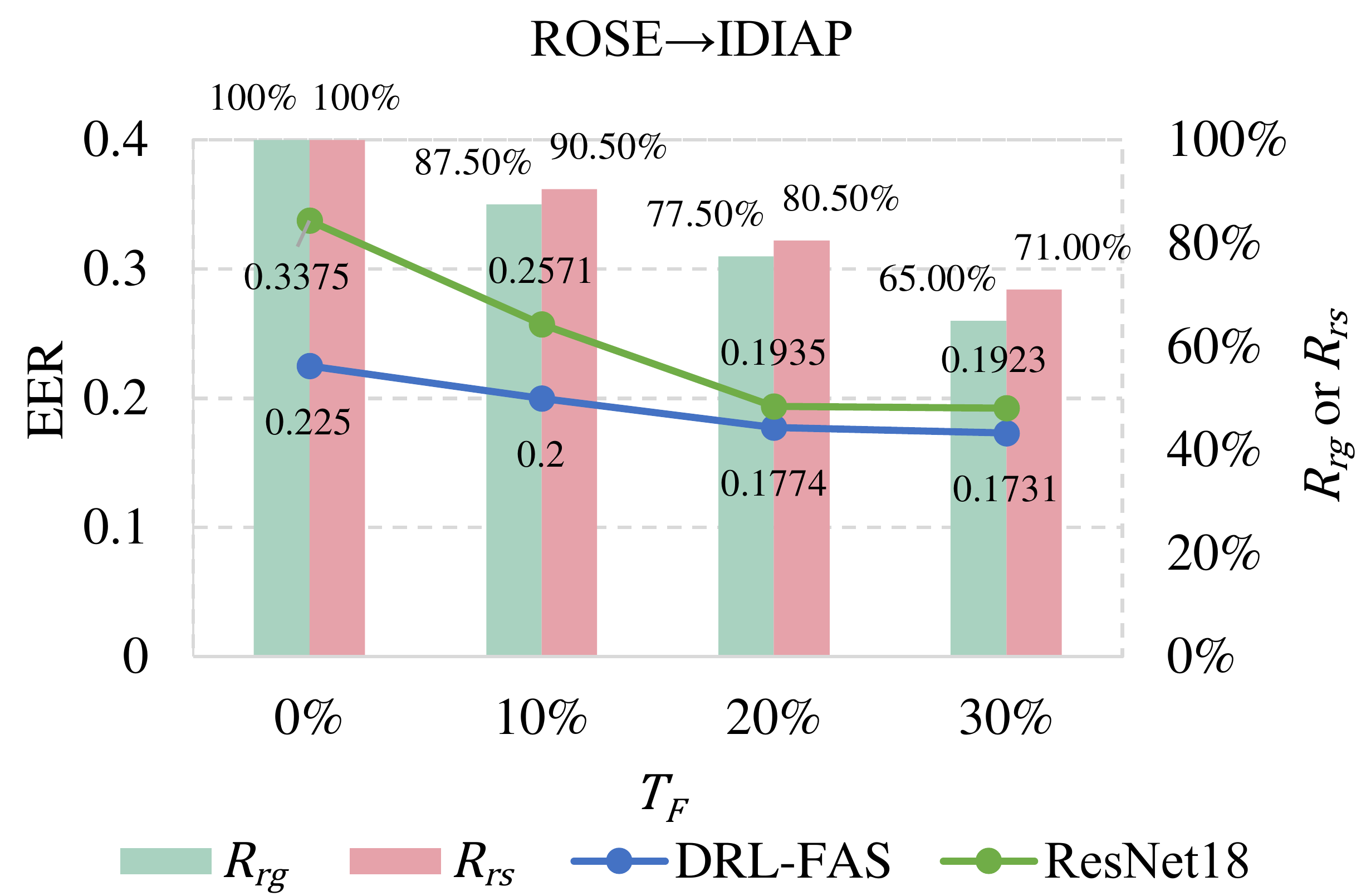}
\hspace{2.5cm}
\includegraphics[width=2.5in]{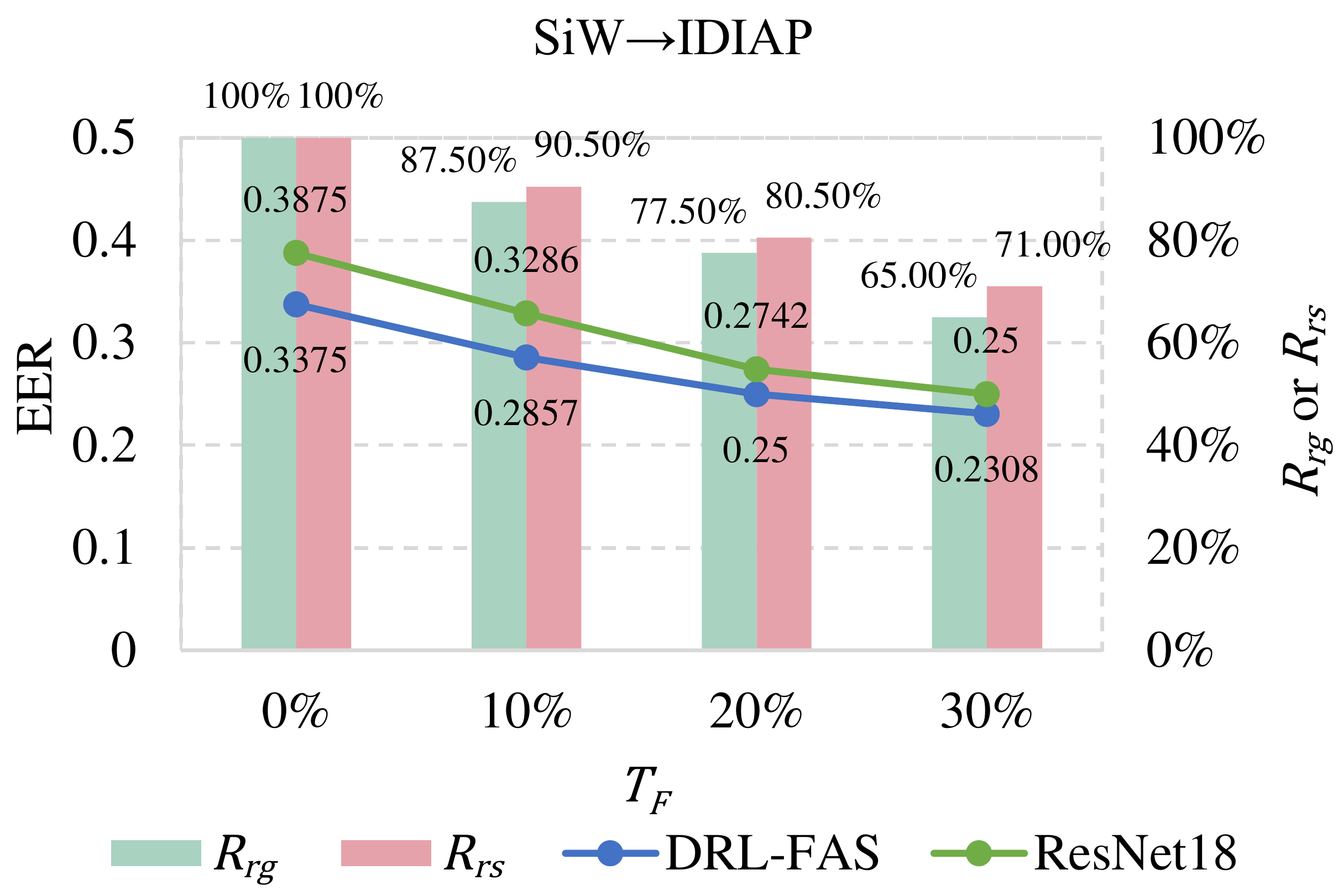}
\end{minipage}
}
\\
\hspace{0.2cm}
\subfigure[w/o unknown attacks]{
\begin{minipage}[t]{0.475\textwidth}
\centering
\includegraphics[width=2.5in]{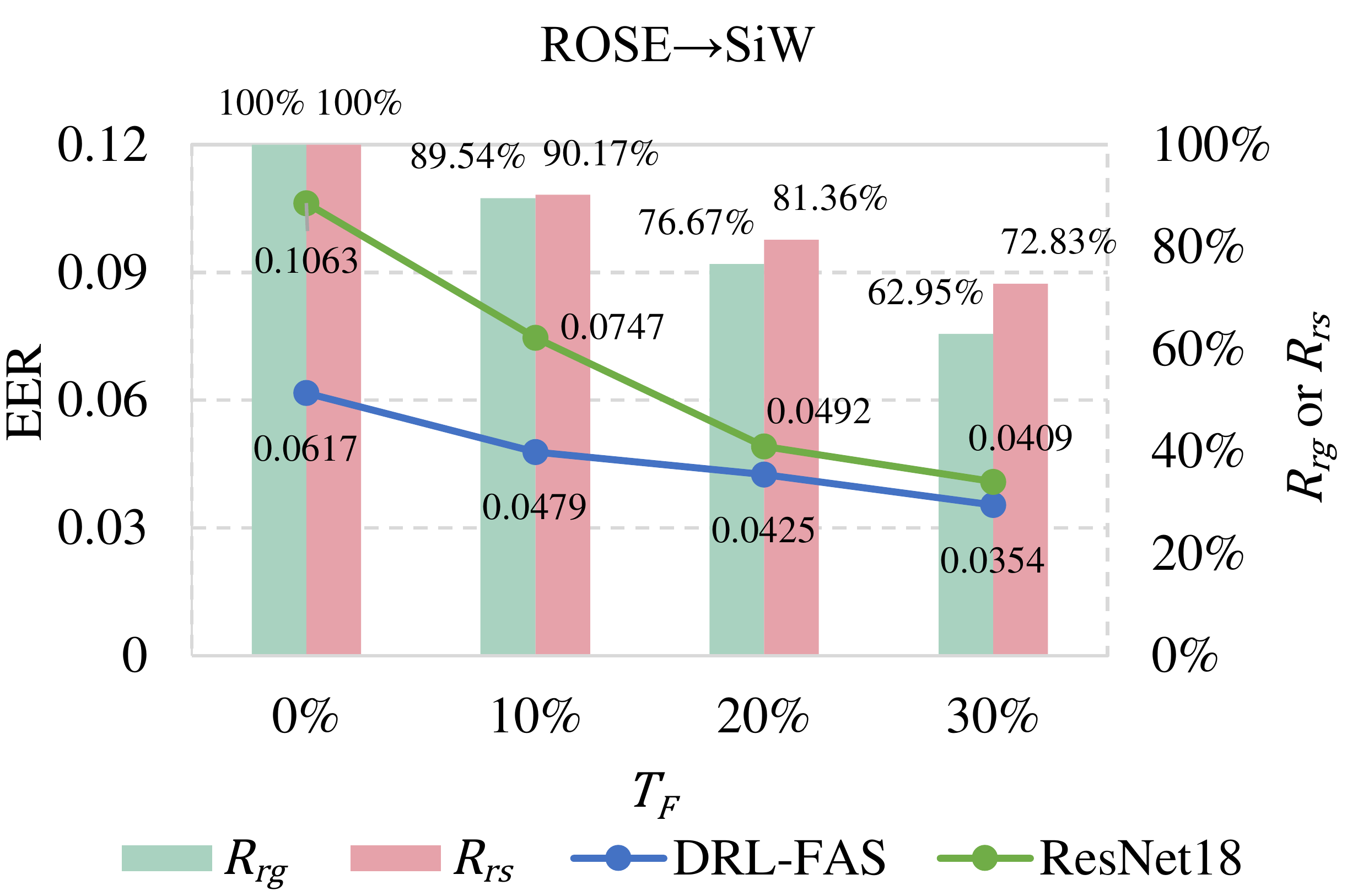}
\end{minipage}
}
\hspace{0.1cm}
\subfigure[with unknown attacks]{
\begin{minipage}[t]{0.475\textwidth}
\centering
\includegraphics[width=2.5in]{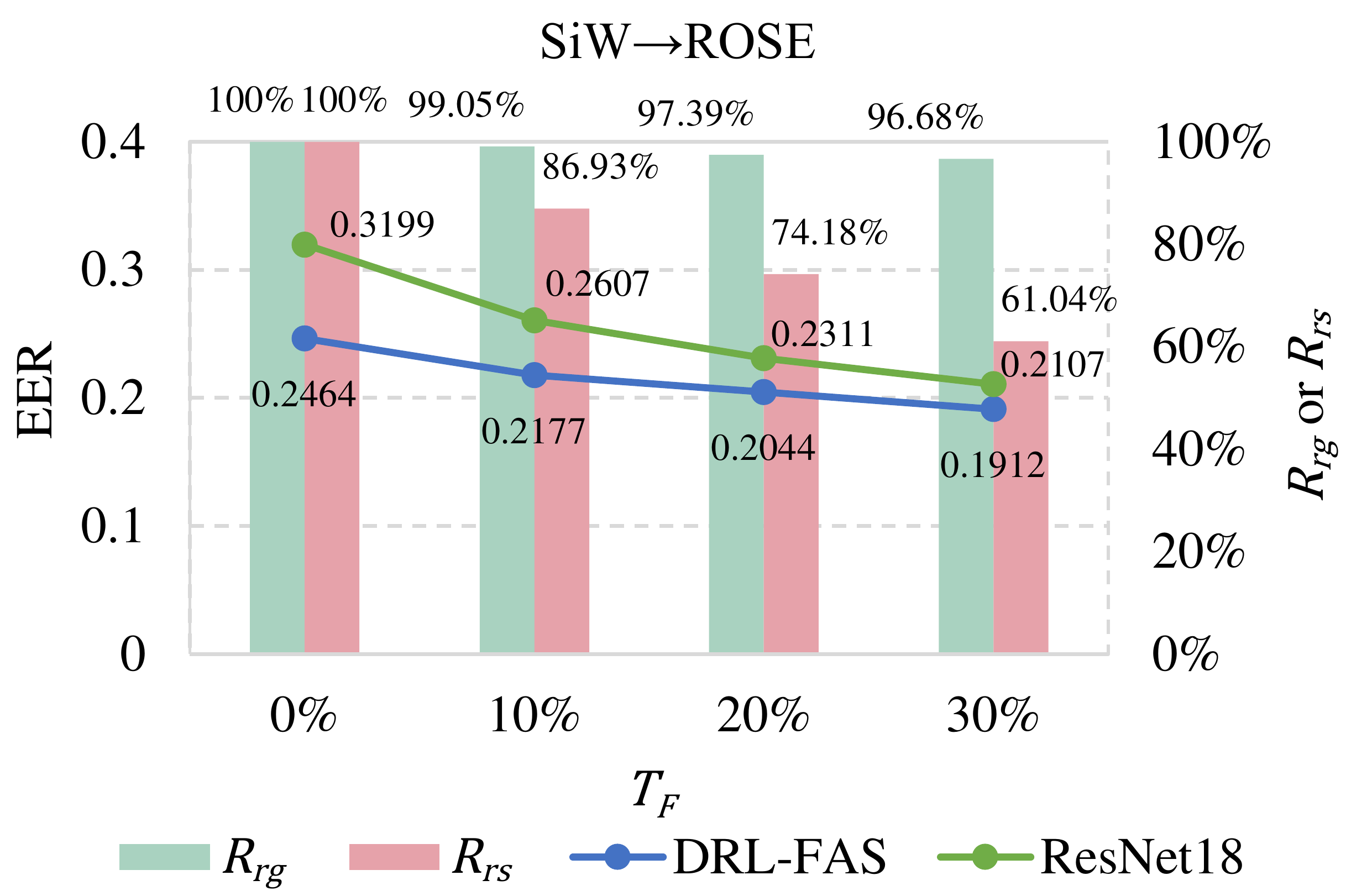}
\end{minipage}
}
\caption{The FAS performances of DRL-FAS \cite{cai2021drl} and ResNet18 \cite{he2016deep} after applying FANet with different thresholds on the testing sets. Four cross-database experiments stand for three representative scenarios. (a) Training on high-quality images and testing on low-quality images. (b) The testing set does not contain samples from unknown attacks. (c) The testing set contains samples from unknown attacks. The bars indicate the ratios of remaining genuine and spoofing samples after filtering, while the curves show the EERs of different approaches in the remaining samples.}\vspace{-0.25cm}
\label{fig:FaceEER}
\end{figure*}

Before the samples are forensically examined, we evaluate the forensicability of these samples through the FANet and reject the samples with low forensicability score, to improve the efficiency of the forensic system. 
In our experiments, we evaluate filtering threshold $T_F \in \{0\%, 10\%, 20\%, 30\% \}$ to simulate forensic scenarios under different security requirements. 
For example, $T_F=0\%$ indicates the configuration without rejecting any samples, and $T_F=10\%$ rejects samples with the lowest 10\% forensics scores.
Specifically, the forensicability assessment process takes a video sequence (i.e., the video frames) as input. 
The forensicability score is calculated for each frame. 
The frame-level scores are averaged to yield the forensicability score of the input video. 
The videos with score lower than $T_F$ are rejected.
Then, we input the remaining samples into the following forensic algorithms for further investigation. 

\subsubsection{Evaluation Metrics}
\label{subsubsec:Evaluation metrics}

In evaluating the proposed FANet, we set filtering threshold as $T_F \in \{0\%, 10\%, 20\%, 30\% \}$. 
The minimum threshold $T_F=0\%$ indicates the condition without filtering, while the maximum value $T_F=30\%$ is set since the improvements have saturated in most protocols between $T_F=20\%$ and 30\%.
We evaluate the filtering results with the ratios of remaining genuine samples and spoofing samples, defined as
\begin{align}
\label{eqn:RemainingRatios}
R_{rg} &= N_{rg}/N_{og},\\
R_{rs} &= N_{rs}/N_{os}
\end{align}
where $N_{rg}, N_{rs}$ denote the number of remaining genuine and spoofing samples after filtering. $N_{og}, N_{os}$ denotes the number of genuine and spoofing samples in the original dataset.
The performance of forensic systems is evaluated in terms of the equal error rate (EER) in the remaining samples after filtering by the FANet. 
The effectiveness of the proposed forensicability framework is then evaluated by comparing the EERs under different configurations of the forensicability filter (i.e., FANet).

\subsection{Experimental Results}
\label{subsec:Results}

\subsubsection{Results on FAS}
\label{subsubsec:Face}

In this section, we choose the generic deep learning-based FAS method \cite{he2016deep} and the SOTA FAS method \cite{cai2021drl} for forensic experiments to evaluate the effectiveness of integrating FANet into a forensic system.

\vspace{0.25cm}
\noindent $\bullet$ \emph{Results of a Generic Deep Learning-based FAS Method}

We first select a deep learning-based forensic method for implementing FANet. 
ResNet \cite{he2016deep} is a generic feature extraction backbone with different layer and depth configurations. 
Among them, we use a feature extraction backbone based on the ResNet18 framework and connect it with the FC layer used for classification. 
The number of input and output channels in the FC layer are 512 and 2, respectively.
In our experiments, we find that the existing forensic methods have achieved good performances in intra-database experiments. 
In order to reflect more practical forensic scenarios in the evaluation of our method, we adopt the cross-database experimental protocol. 
Specifically, the ResNet18-based classification network is trained and tested with different datasets. 
Four cross-database experiments are conducted, ROSE→IDIAP, ROSE→SiW, SiW→IDIAP, and SiW→ROSE, where the datasets on the left and right sides of the arrow indicate the training and testing sets, respectively.

We set 3 filtering thresholds to reject the samples with the lowest 10\%, 20\%, and 30\% forensic scores in the testing set.
As illustrated by the bars in Fig.~\ref{fig:FaceEER}, different ratios of genuine and spoofing samples are rejected in these datasets.
It is mainly because the samples in these datasets are of different qualities.
In the protocols where IDIAP and SiW are employed as the testing set, $R_{rg}$ decreases faster than $R_{rs}$ as $T_F$ increases.
In these cases, the samples in both datasets are generally of lower quality than those in the training dataset of FANet (i.e., CASIA).
Therefore, the FANet tends to reject a larger percentage of low-forensicability samples in the genuine samples (of higher quality than the spoofing samples).
Meanwhile, in the protocols where the high-quality ROSE dataset is employed as the testing set, $R_{rs}$ decreases faster than $R_{rg}$ as $T_F$ increases.
It can be explained by the same reason as above.



As depicted by the curves in Fig.~\ref{fig:FaceEER}, we report the EERs of different configurations in four cross-database protocols. 
Depending on the image quality and types of attack in the datasets, the four cross-database experiments evaluate the system performance under three different scenarios. 
ROSE→IDIAP and SiW→IDIAP represent scenarios trained with high-quality (HQ) dataset and tested with low-quality (LQ) dataset. 
In the HQ→LQ scenario, the ResNet18-based approach has achieved 33.75\% and 38.7\% EER under ROSE→IDIAP and SiW→IDIAP protocols, respectively. 
It can be seen from Fig.~\ref{fig:FaceEER} that the performance of the forensic system is significantly improved with FANet. 
In the experiment of ROSE→IDIAP, when the filtering threshold $T_F=10\%$, the EER is reduced by 8.04 percentage points (p.p.) compared to that of the original configuration with FANet (i.e., $T_F=0\%$), which means a 23.82\% performance improvement. 
When $T_F=30\%$, the EER is reduced by 14.52 p.p., which is a 43.02\% improvement over that of no FANet (i.e., $T_F=0\%$). 
A similar conclusion can be found in the experiment of SiW→IDIAP. However, the improvement is not as significant as that under protocol ROSE→IDIAP. 
This is because SiW dataset is less diverse than ROSE dataset, and the generalization performance of the model trained by SiW dataset is weaker than that by the ROSE dataset. 

Protocols ROSE→SiW and SiW→ROSE show the performances under the scenarios without and with unknown attacks, respectively.
The spoofing faces in the ROSE database are from print attack, replay attack, and mask attack, while those in the SiW database only covers the first two types of attack. 
For the model trained with the SiW database, mask attack is an unknown attack. 
Therefore, the FAS performance of SiW→ROSE is significantly weaker than that of ROSE→SiW. 
Our FANet leads to effective forensic system performance under both scenarios with or without unknown spoofing attacks. 
The ResNet18-based method achieves 10.63\% EER in experimental protocol ROSE→SiW (without unknown attack), while it achieves 31.99\% EER in protocol SiW→ROSE (with unknown attack).
For both ROSE→SiW and SiW→ROSE protocols, the EERs are reduced by 6.54 p.p. and 10.92 p.p. after applying filtering threshold $T_F=30\%$. 
In other words, the system performance has been significantly improved by 61.52\% and 34.14\%.

Moreover, we investigate whether the samples rejected by FANet are indeed difficult to be classified by the forensic network.
In this study, we select the most difficult protocol, ROSE→SiW, and report the EER of samples rejected by FANet.
Compared with other protocols, this protocol has the lowest EER, 10.63\% at $T_F=0\%$.
So it is more difficult to improve the performance.
However, the quantitative results in Tab.~\ref{tab:EERofFiltered} show that the EERs of rejected samples under different $T_F$ are high.
The EERs of samples filtered by $T_F = 10\%, 20\%$ and 30\% are 80.33\%, 69.85\% and 56.48\%, respectively.
The qualitative results are shown in Fig.~\ref{fig:FaceLowForensicability}.
Some of the filtered samples are of poor image quality, while others have significant anomalies in terms of color or noise.
Thus, the rejected samples are indeed difficult to be classified by the ResNet18-based forensic network, which demonstrates that the application of FANet is effective and reasonable.

\begin{table}[t!]
\centering
\caption{The EERs of different FAS methods in the samples rejected by the proposed FANet under ROSE→SiW protocol.}
\label{tab:EERofFiltered}
\begin{tabular}{|c|c|c|c|}
\hline
\backslashbox{Methods}{Threshold $T_F$}  & 10\%   & 20\%   & 30\%   \\ \hline
ResNet18 \cite{he2016deep}     & 0.8033 & 0.6985 & 0.5648 \\ \hline
DRL-FAS \cite{cai2021drl}      & 0.6885 & 0.5515 & 0.4537 \\ \hline
\end{tabular}
\vspace{-0.25cm}
\end{table}

Finally, we perform ablation studies to investigate the importance of forensic features in FANet by rejecting the same amount of low-forensicability samples under different configurations. 
The studied configurations include: 1) training FANet without low forensicability class, 2) training FANet by random center initialization without using forensic domain knowledge, and 3) training FANet without the forensic features. 
It is noted that the trained latent spaces of the first two settings have also been presented in Fig.~\ref{fig:ForensicabilityDistribution}, and the third setting can be considered as simple extensions from methods that employ image quality features in FAS task \cite{li2016face, fourati2017face, fourati2020anti}.
Here, we use a ResNet18-based FAS network trained with the ROSE training set, and evaluate the system performance on the IDIAP testing set.
The compared models trained by omitting important pieces from the proposed FANet show a significant deterioration in EERs.
Our FANet shows 19.35\% EER under $T_F=20\%$, which is a 42.67\% improvement over that of no FANet (i.e., $T_F=0\%$). 
The configurations without low-forensicability class, without using forensic domain knowledge in center initialization, and without using forensic features lead to EERs of 24.19\%, 33.87\% and 30.61\%, respectively, which are 25.01\%, 75.04\% and 58.19\% worse than that of the proposed FANet.
These ablation studies demonstrate the importance of introducing the low forensicability class, initializing centers with forensic domain knowledge, and extracting forensic features in FANet.

\begin{figure}[t!]
\centering
\subfigure[]{
\begin{minipage}[t]{0.45\linewidth}
\centering
\includegraphics[width=1.5in]{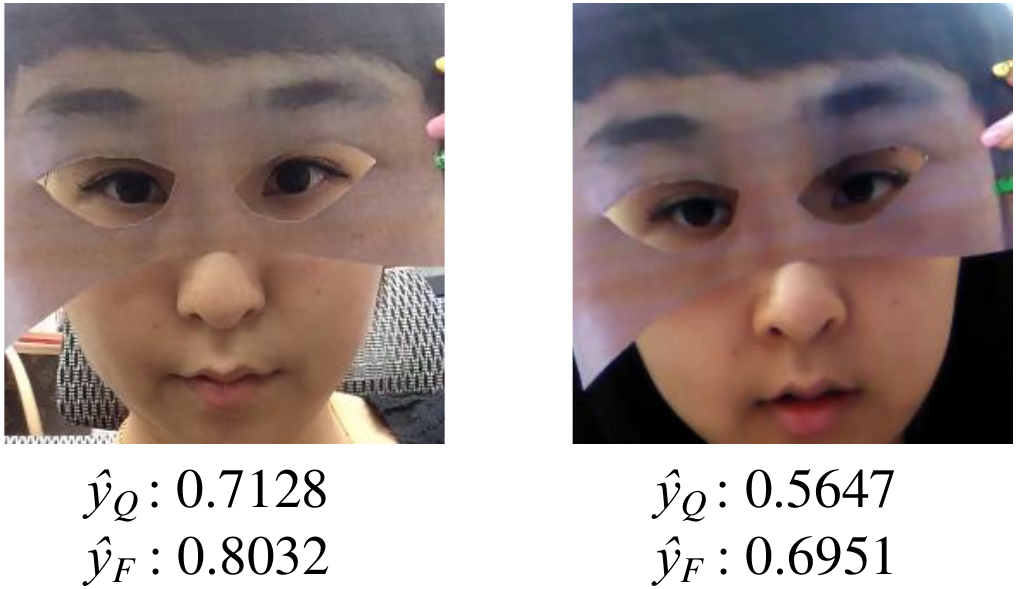}
\end{minipage}
}
\enspace
\subfigure[]{
\begin{minipage}[t]{0.45\linewidth}
\centering
\includegraphics[width=1.5in]{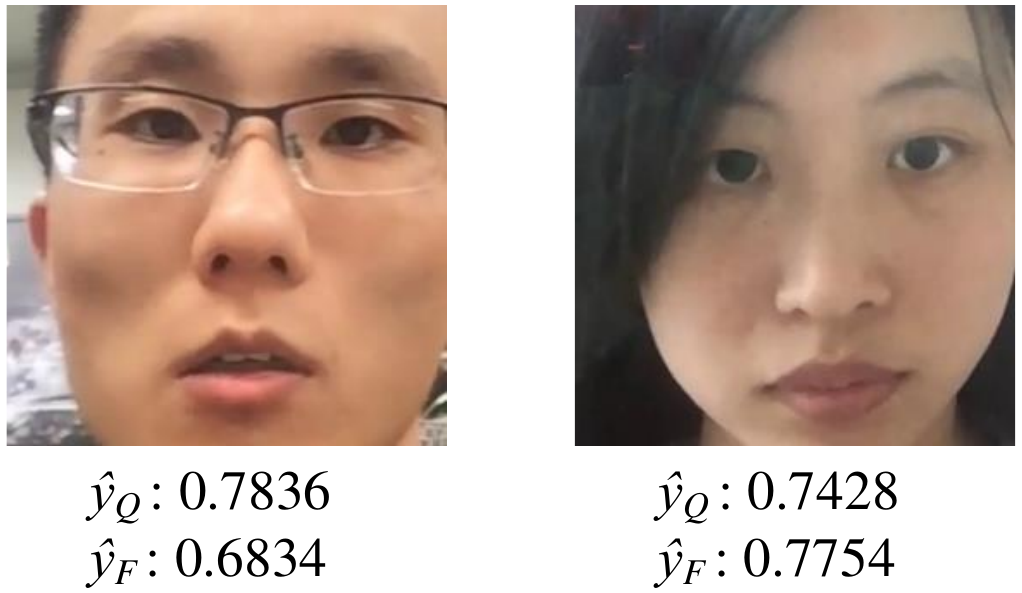}
\end{minipage}
}
\vspace{-0.25cm}
\caption{Error samples under the SiW→ROS protocol in the high-forensicability samples pass the FANet. $\hat{y}_Q$ denotes image quality score and $\hat{y}_{F}$ denotes forensic score. (a) Samples from unknown attacks. (b) High-quality recapturing samples.}\vspace{-0.25cm}
\label{fig:FaceErrorSamples}
\end{figure}

\vspace{0.25cm}
\noindent $\bullet$ \emph{Results of a State-of-the-art FAS Method}



As demonstrated in the previous section, the generic deep learning-based FAS method does not perform well in cross-database experiments. 
In recent years, the SOTA FAS methods have shown significant performance improvements in cross-database scenarios. 
In order to evaluate the performance of FANet in a SOTA security system, we select a SOTA FAS method for our experiment with the following criteria. 
First, both the source code and weights of trained model should be available to facilitate comparative experiments. 
Second, the architecture of the method should be significantly different from the generic CNN architecture to yield more representative results.
Third, since we assess the forensicability of samples from RGB data, the selected method preferably does not require additional supervision information, e.g., depth information. 
Thus, we select the DRL-FAS \cite{cai2021drl} method for our experiments. 
In our experiments, when training the DRL-FAS network, we follow the original setup in \cite{cai2021drl} to ensure a fair comparison.


As shown in Fig.~\ref{fig:FaceEER}, the DRL-FAS model achieves 22.50\% and 6.17\% EERs in the experimental protocols of ROSE→IDIAP and ROSE→SiW, respectively. 
Then, we apply different filtering thresholds to the testing set by FANet. 
After filtering with $T_F=$10\%, the EERs are decreased by 2.50 p.p. and 1.38 p.p., respectively.
Under $T_F=$30\%, the EERs are decreased by 5.19 p.p. and 2.63 p.p., respectively.
It shows that our method with simple CNN-based forensic features is effective in identifying low forensicability samples for the SOTA FAS approach. 
Similarly, in the experiments of SiW→IDIAP and SiW→ROSE, the performances are improved by 10.67 p.p. and 5.52 p.p., respectively, after filtering with $T_F=30\%$.

We also analyze the samples rejected by FANet.
Following the previous section, we select the protocol ROSE→SiW to perform the experiment.
The quantitative results are shown in Tab.~\ref{tab:EERofFiltered}.
Under thresholds $T_F=$10\%, 20\% and 30\%, the EERs of DRL-FAS methods in the rejected samples are 68.85\%, 55.15\%, and 45.37\%, respectively.
These quantitative results illustrate that even the state-of-the-art FAS method performs poorly in the low-forensicability samples rejected by the proposed FANet.
FANet effectively filters these samples and improves the efficiency of the forensic system.

Admitably, our FANet in its current form only addresses a portion of the face spoofing problems. 
As shown in Fig.~\ref{fig:FaceLowForensicability} and Fig.~\ref{fig:FaceErrorSamples}, FANet does not work well for spoofing samples with high image quality or from unknown types of attack. 
However, as the experimental results shown in the challenging protocol SiW→ROSE (with unknown attack), the proposed FANet filters the samples with low forensicability, thus improving the efficiency of the forensic system. 
Moreover, our method is also applicable to the samples from different types of attack given that more diverse training samples (besides the synthesis samples in Sec.~\ref{subsec:Settings}) are employed in training our FANet.

\vspace{0.25cm}
\noindent $\bullet$ \emph{Application in Practical Scenes}

\begin{figure}[t!]
\subfigure[]{
\begin{minipage}[t]{0.475\linewidth}
\centering
\includegraphics[width=1.8in]{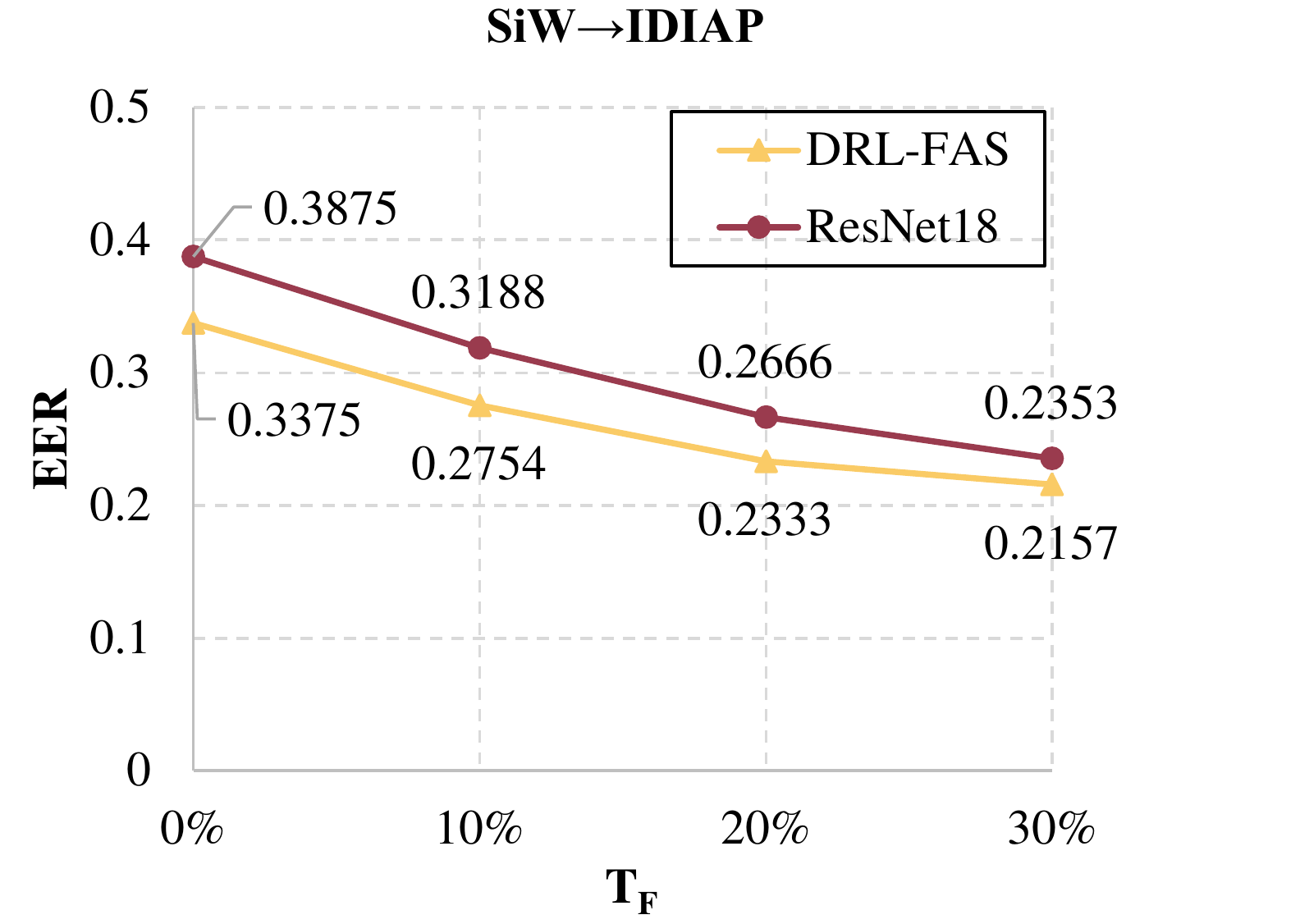}
\end{minipage}
}
\hspace{-0.25cm}
\subfigure[]{
\begin{minipage}[t]{0.475\linewidth}
\centering
\includegraphics[width=1.8in]{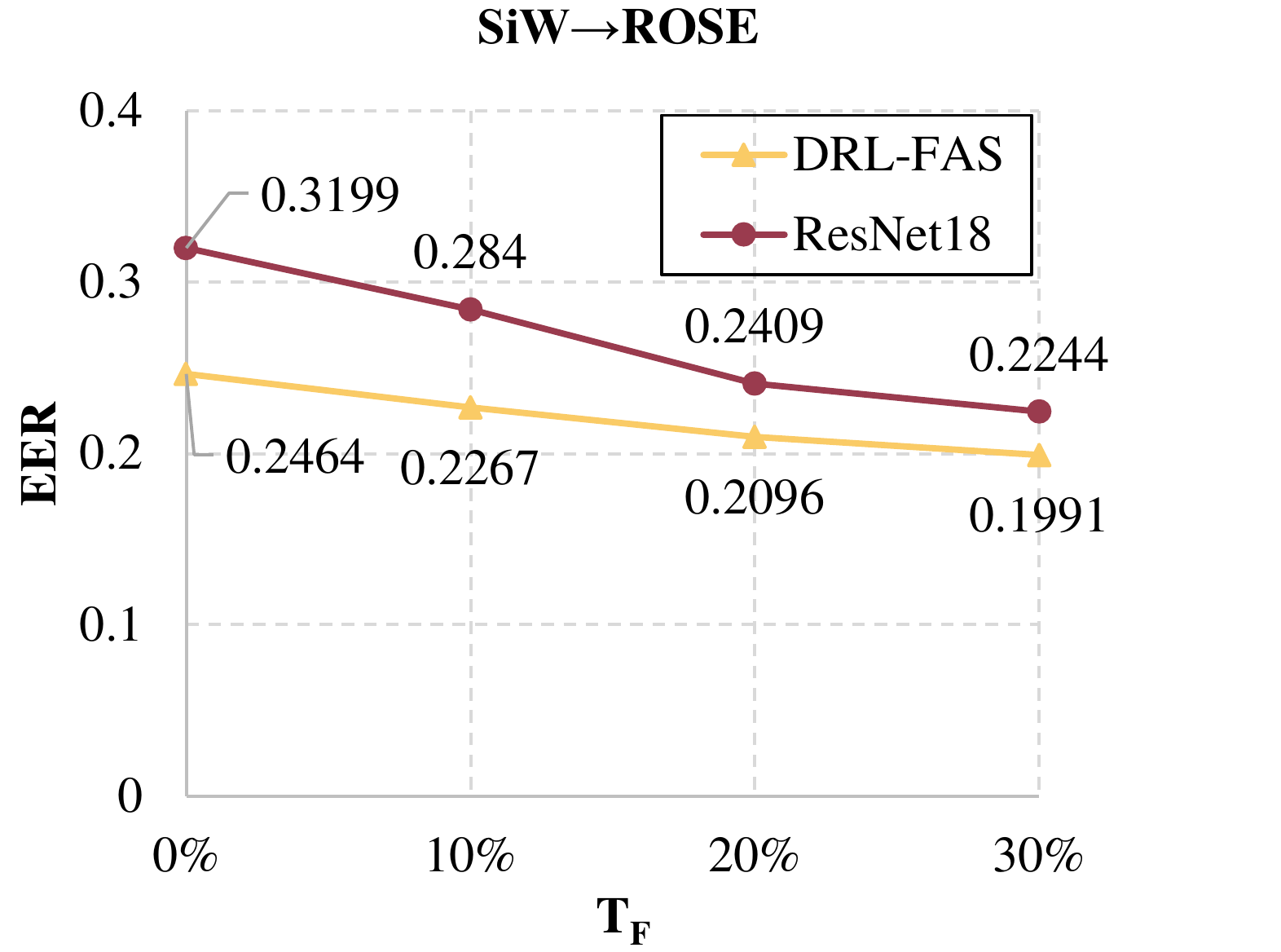}
\end{minipage}
}
\vspace{-0.5cm}
\caption{The performances of DRL-FAS \cite{cai2021drl} and ResNet18 \cite{he2016deep} after applying FANet with different thresholds. We employ forensic models trained by the training set of the SiW database. The forensicability thresholds are defined through the testing set of the SiW database. (a) SiW→IDIAP: testing on low-quality images. (b) SiW→ROSE: testing on high-quality images.}\vspace{-0.25cm}
\label{fig:PracticalEER}
\end{figure}

In the results presented in Fig.~\ref{fig:FaceEER}, different thresholds (determined by forensicability scores of the testing sets) are employed across different experimental protocols even under the same setting of $T_F$.
Considering the application of FANet in a more practical scenario, we attempt to fix the threshold on forensicability scores with the help of a publicly available dataset, and then conduct cross-dataset experiments based on these forensicability scores under different settings of $T_F$.
The specific implementation steps are as follows.
Firstly, we compute the forensicability scores of the testing set in SiW database with FANet, and the corresponding thresholds on forensicability scores are determined according to different settings of $T_F \in \{10\%, 20\%, 30\% \}$.
Secondly, we train the FAS networks (both ResNet18 and DRL-FAS) with the training set of SiW dataset.
Thirdly, we perform cross-dataset experiments with the corresponding thresholds on forensicability scores under $T_F = 10\%, 20\%$, and $30\%$.
Based on the output forensicability scores, the testing set is filtered and the performances of the FAS networks are verified under protocols SiW→IDIAP and SiW→ROSE.

The results of the cross-database experiments with fix threshold on forensicability scores are shown in Fig.~\ref{fig:PracticalEER}.
The proposed FANet is effective since the performances of the forensic models are continuously improved with the increment of $T_F$.
Among two databases, the IDIAP database is of lower quality.
When the same threshold on forensicability score is applied across different protocols, more samples of IDIAP are rejected under the same forensicability score.
Therefore, compared to Fig.~\ref{fig:FaceEER}, the performance improvement under SiW→IDIAP is more significant under this experiment, while and that under SiW→ROSE is less significant.


\vspace{0.25cm}
\noindent $\bullet$ \emph{Computational Complexity}


\begin{table}[t!]
\centering
\caption{Computational complexity and performance of the proposed FANet with different FAS approaches. The experimental results are obtained under protocol ROSE→IDIAP.}
\label{tab:ComputationalComplexity}
\vspace{-0.25cm}
\begin{tabular}{|c|c|c|ccc|}
\hline
\multirow{2}{*}{Methods}                                                   & \multirow{2}{*}{Metrics}   & \multirow{2}{*}{w/o FANet}                                       & \multicolumn{3}{c|}{FANet with different $T_F$}                                                              \\ \cline{4-6} 
                                                                           &                                                             &     & \multicolumn{1}{c|}{10\%}   & \multicolumn{1}{c|}{20\%}   & 30\%   \\ \hline
\multirow{2}{*}{\begin{tabular}[c]{@{}c@{}}FANet\\ +ResNet18\end{tabular}} & G-FLOPs                                                     & \multicolumn{1}{c|}{1.82}   & \multicolumn{1}{c|}{2.54}   & \multicolumn{1}{c|}{2.36}   & 2.17   \\ \cline{2-6} 
                                                                           & EER & \multicolumn{1}{c|}{0.3375} & \multicolumn{1}{c|}{0.2571} & \multicolumn{1}{c|}{0.1935} & 0.1923 \\ \hline
\multirow{2}{*}{\begin{tabular}[c]{@{}c@{}}FANet\\ +DRL-FAS\end{tabular}}  & G-FLOPs                                                     & \multicolumn{1}{c|}{2.41}   & \multicolumn{1}{c|}{3.07}   & \multicolumn{1}{c|}{2.83}   & 2.59   \\ \cline{2-6} 
                                                                           & EER & \multicolumn{1}{c|}{0.2250} & \multicolumn{1}{c|}{0.2000} & \multicolumn{1}{c|}{0.1774} & 0.1731 \\ \hline
\end{tabular}
\end{table}

Since our proposed method involves a two-step operation, i.e., the FANet and the forensic network, we analyze the computational complexity of each step in the FANet+ResNet18 and the FANet+DRL-FAS systems. 
The system complexity is computed by
\begin{align}
\label{eqn:Complexity}
\begin{split}
O_S = O_{FA} + (1 - T_F) \cdot O_{FI},
\end{split}
\end{align}
where $O_{FA}$, $O_{FI}$ and $O_S$ denote the number of operations of FANet, forensic inspection, and the whole system, respectively. 
As shown in Tab.~\ref{tab:ComputationalComplexity}, the computational complexity (measured in G-FLOPs) of the system is lower as the threshold increases since more low-forensicability samples are rejected without going through the forensic inspection process. 
We also note that after 30\% filtering, the computational complexities of both systems are already close to the version without FANet. 
For the FANet+ResNet18 method under $T_F=30\%$, although the complexity is increased by 19.23\% over the setting without FANet, it is in exchange for 43\% performance improvement in EER under the cross-database protocol ROSE→IDIAP. 
As for the DRL-FAS method, it is with a higher computational complexity compared to ResNet18. 
After rejecting 30\% low-forensicability samples, the complexity of the FANet+DRL-FAS system has only been increased by 7.5\% compared to that of the original forensic network while the EER has been reduced by 29.98\%.

\begin{figure}[t!]
\centering
\subfigure[Captured student ID images.]{
\hspace{-0.25cm}
\begin{minipage}[t]{1\linewidth}
\centering
\includegraphics[width=1.1in]{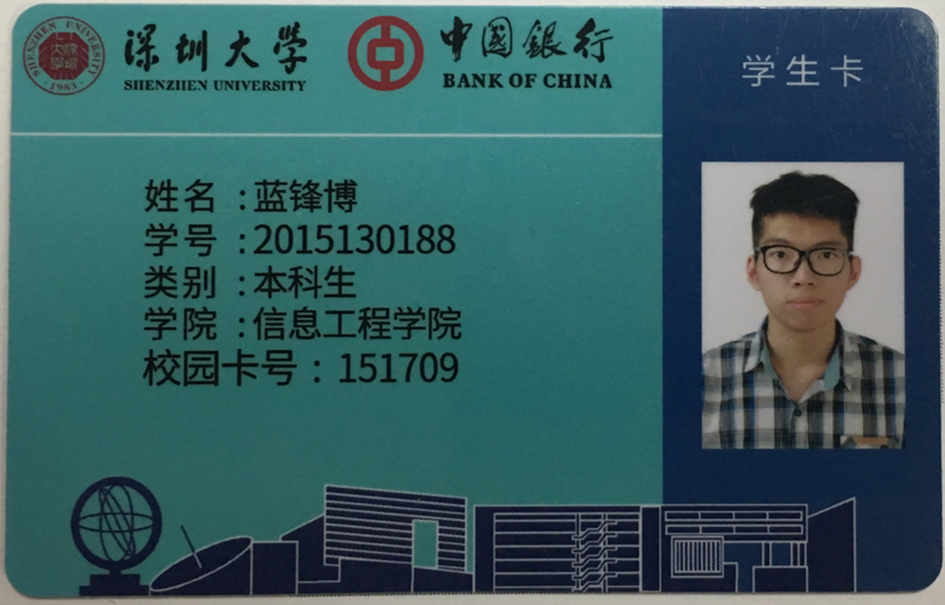}
\includegraphics[width=1.1in]{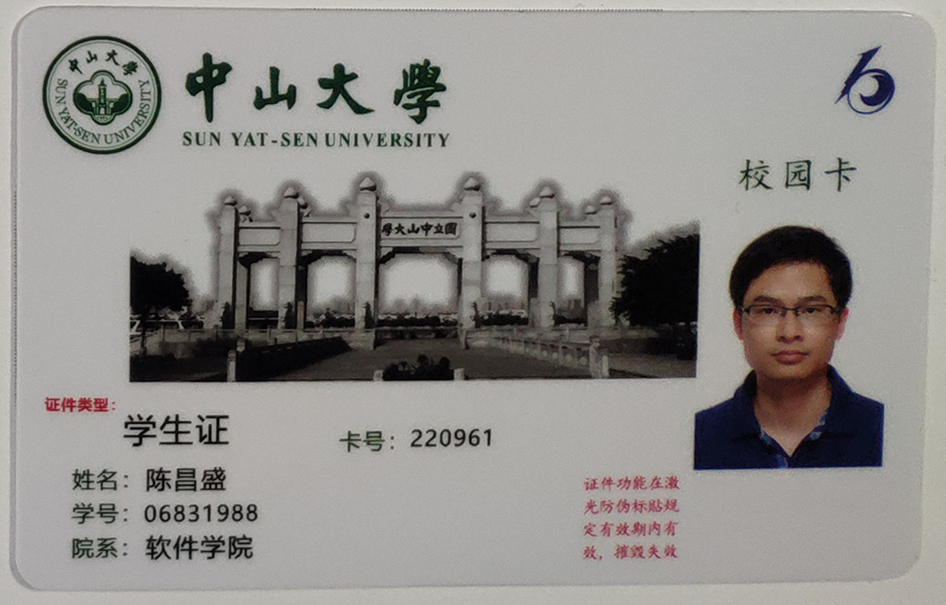}
\includegraphics[width=1.1in]{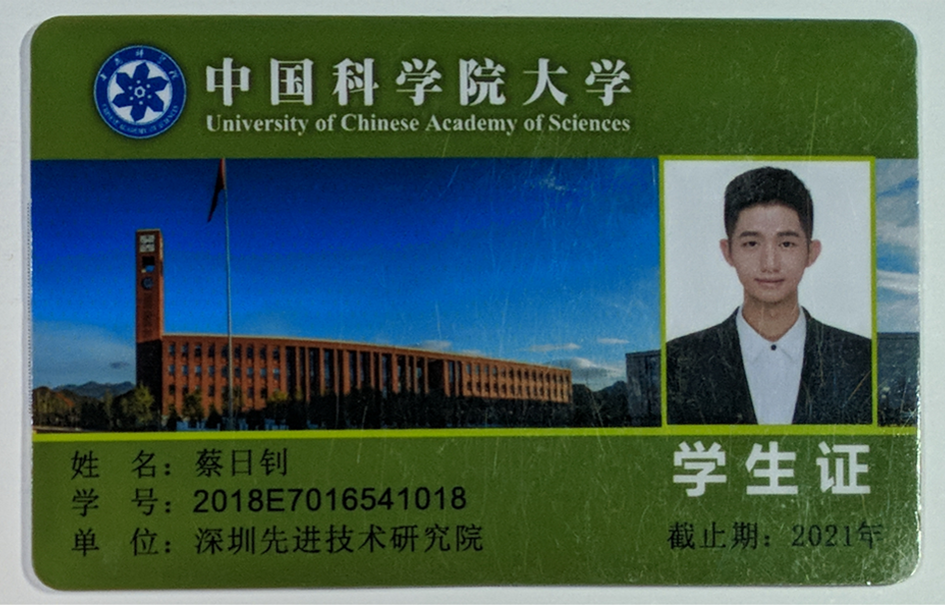}
\end{minipage}
}
\subfigure[Recaptured student ID images.]{
\hspace{-0.25cm}
\begin{minipage}[t]{1\linewidth}
\centering
\includegraphics[width=1.1in]{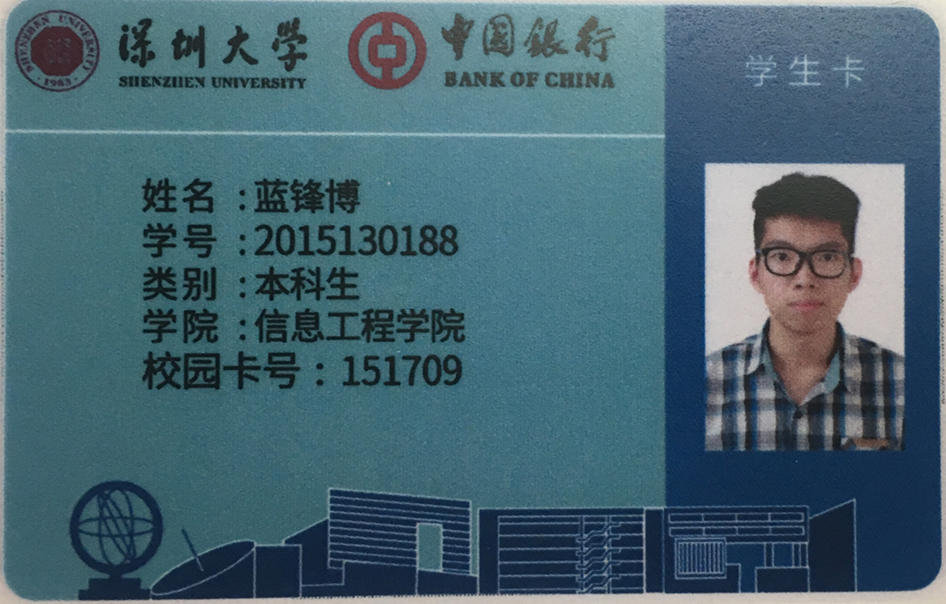}
\includegraphics[width=1.1in]{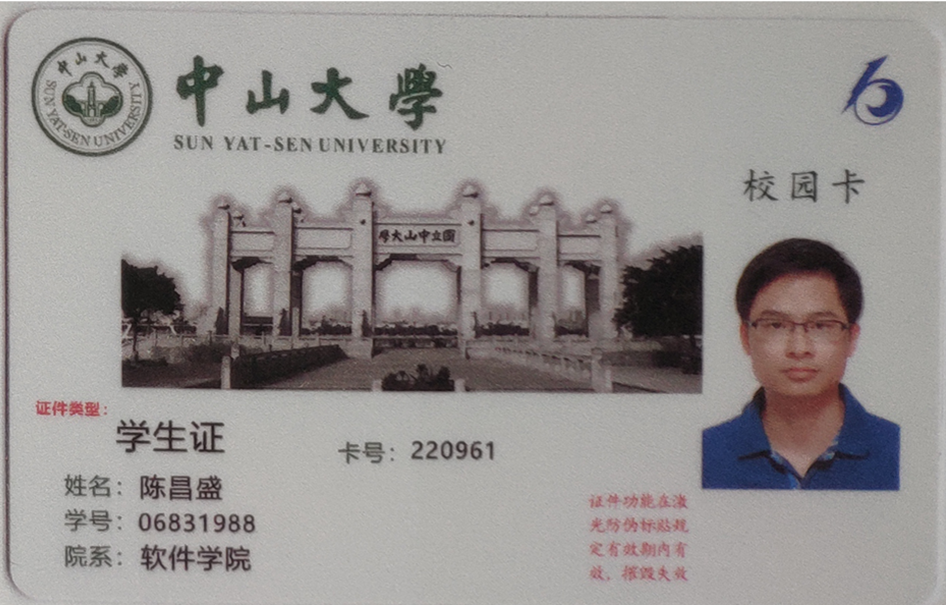}
\includegraphics[width=1.1in]{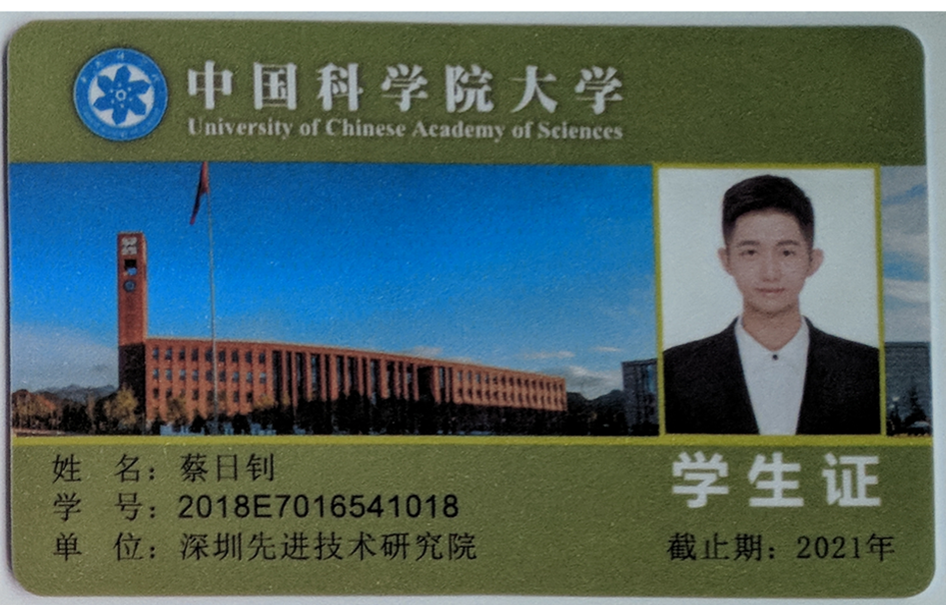}
\end{minipage}
}
\caption{Comparison of captured and recaptured student ID images in \cite{chen2021domain}.}\vspace{-0.25cm}
\label{fig:DocumentSubjective}
\end{figure}

\vspace{0.25cm}
\subsubsection{Results on Recaptured Document Detection}
\label{subsubsec:Document}



To further evaluate the generalization performance of our proposed method, we select the recaptured document detection task for our experiments, whose purpose is to classify captured and recaptured documents. 
According to \cite{zhao2021deep}, detecting recaptured documents is very different from detecting spoofing in face images. 
Moir\'{e} pattern artifacts and other forensic traces from depth in the face images are not available in the document images. 
Both the captured and recaptured versions of the hardcopy documents are acquired from a flat paper surface, which lacks the differences in texture patterns and depth. 
Therefore, it is a challenging forensic task. 

For the recaptured document detection task, we follow the experiments on FAS in Sec.~\ref{subsubsec:Face} for selecting forensic methods. 
First, we select the ResNet18 which is the same as the generic deep learning approach in our FAS experiments.
Second, we also employ a SOTA recaptured document forensic network proposed in \cite{chen2021domain} with ResNet50 backbone, i.e., ResNet50+FS\&TL.

\begin{figure}[t!]
\vspace{0.25cm}
\centerline{\includegraphics[width=0.75\linewidth]{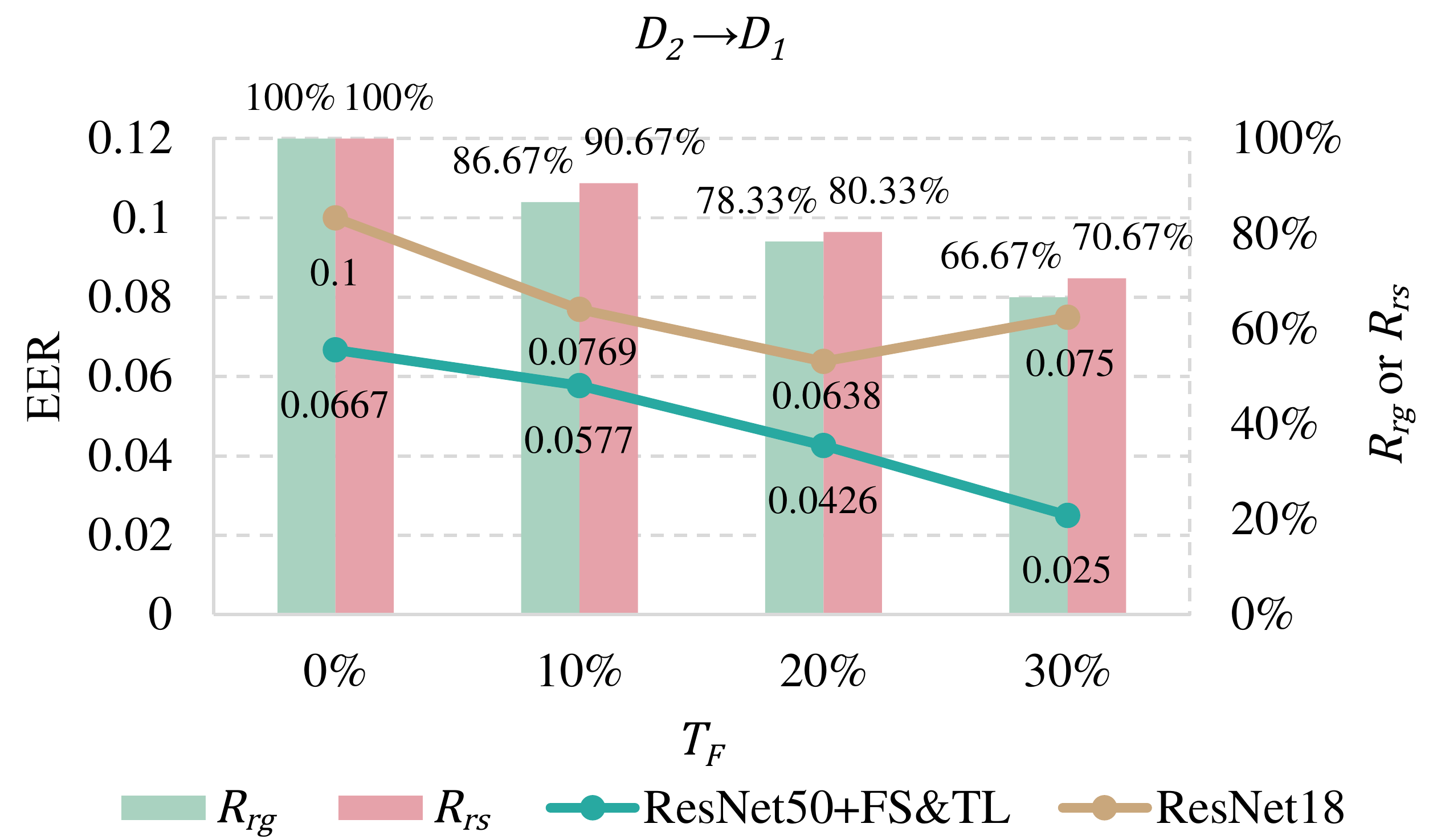}}
\vspace{-0.25cm}
\caption{The recaptured document detection performances of ResNet50 +FS\&TL \cite{chen2021domain} and ResNet18 \cite{he2016deep} after applying FANet with different thresholds on the testing sets.}\vspace{-0.25cm}
\label{fig:DocumentEER}
\end{figure}

Employing the recaptured document dataset in \cite{chen2021domain}, we conduct a cross-database experiment with protocol $D_2$→$D_1$. 
As shown in Fig.~\ref{fig:DocumentEER}, ResNet18 and ResNet+FS\&TL \cite{chen2021domain} achieved 10.00\% EER and 6.67\% EER, respectively, before implementing FANet ($T_F=0\%$).
Then, we apply FANet to reject the low-forensicability samples in $D_1$. 
The statistics of filtered samples are shown by the bars in Fig.~\ref{fig:DocumentEER}. 
We note that as $T_F$ increases, $R_{rg}$ decreases faster than $R_{rs}$.
The samples in $D_2$ is of higher quality than those in dataset $D_1$ \cite{chen2021domain}.
For the same reason mentioned in Sec.~\ref{subsubsec:Face}, FANet filters out more samples with low forensicability in the captured (or genuine) images.
After applying the 30\% threshold, the EERs of ResNet18 and ResNet50+FS\&TL \cite{chen2021domain} are reduced by 2.50 p.p. and 4.17 p.p., respectively. 
These show performance improvement by 25.00\% and 62.52\%, respectively. 
It indicates that FANet is also effective on the recaptured document detection task. 
In addition, we analyze the performance of the samples filtered by FANet in the ResNet50+FS\&TL method.
The EERs of filtered samples are 62.50\%, 53.85\% and 40.00\% under $T_F$ = 10\%, 20\% and 30\%, respectively.
The results show that the samples with low forensicability filtered by FANet are prone to misclassification by the forensic networks.
Our FANet improves the reliability of the forensic system.
We also note that after applying the 30\% filtering threshold, most of the remaining error samples are high-quality forgery samples. 










\section{Conclusion}
\label{sec:Conclusion}


In this work, we have proposed the FANet to evaluate the forensicability of questioned samples, reject the samples with weak forensic cues, and improve the efficiency of a forensic system. 
The performance of FANet is evaluated under recapturing detection tasks.
Experimental results show that our FANet reduces over 1/3 of the classification EERs of some generic CNNs by rejecting 30\% low-forensicability samples. 
The EER of the rejected samples is higher than 50\% in both FAS and document recapturing detection tasks.
Analysis of computational complexity has also demonstrated that our FANet improves the performance of a SOTA FAS method without causing a significant additional computational burden.
Moreover, the forensicability assessment framework is also extendable to other important forensic tasks.
For example, the detection of different image post-processing operations, such as double JPEG compression, filtering are similar to the detection of recaptured images.

In the future, we will explore the proposed framework with a richer collection of training samples, besides those generated by a simple synthesis method from the CASIA dataset.
A large and diverse training set would lead to better system performance, especially in the case of unknown attacks.

\ifCLASSOPTIONcaptionsoff
  \newpage
\fi

\bibliographystyle{IEEEtran}
\bibliography{References}

\end{document}